\documentclass[11pt,a4paper]{article}
\usepackage[hyperref]{emnlp2020}
\usepackage{times}
\usepackage{latexsym}

\usepackage{amsmath,amsfonts,amssymb}
\DeclareMathOperator*{\argmax}{arg\,max}

\usepackage{enumitem}
\usepackage{graphicx}
\usepackage{booktabs}
\usepackage{subcaption}

\usepackage{tikz}
\usetikzlibrary{arrows,positioning,automata,calc,shapes}
\usepackage{pgfplots}
\pgfplotsset{compat=newest, scaled z ticks=false} 
\pgfplotsset{plot coordinates/math parser=false}
\newlength\figureheight 
\newlength\figurewidth

\definecolor{txtcolor1}{rgb}{1,0.65,0}%
\definecolor{txtcolor2}{rgb}{1,0.94,0}%
\definecolor{txtcolor3}{rgb}{1.0, 0.25, 0.25}%
\definecolor{txtcolor4}{rgb}{0.39, 0.58, 0.93}
\definecolor{txtcolor5}{rgb}{0., 0.18, 0.39}
\definecolor{txtcolor6}{rgb}{0.01, 0.75, 0.24}
\newcommand{\vagrad}{%
        \begin{tikzpicture}[inner sep=0pt, baseline=(base)]%
        \draw[txtcolor1,line width=1.5pt](0,0) -- (5mm,0); 
        \node[very thick, mark size=3pt,color=txtcolor1] at (2.5mm,0){% 
            \pgfuseplotmark{asterisk}%
        };
        \node (base) at (0,-.5ex) {};
        \end{tikzpicture}%
}
\newcommand{\smoothgrad}{%
        \begin{tikzpicture}[inner sep=0pt, baseline=(base)]%
        \draw[txtcolor2,line width=1.5pt](0,0) -- (5mm,0); 
        \node[very thick, mark size=3pt,color=txtcolor2] at (2.5mm,0){% 
            \pgfuseplotmark{triangle}%
        };
        \node (base) at (0,-.5ex) {};
        \end{tikzpicture}%
}
\newcommand{\itergrad}{%
        \begin{tikzpicture}[inner sep=0pt, baseline=(base)]%
        \draw[txtcolor3,line width=1.5pt](0,0) -- (5mm,0); 
        \node[very thick, mark size=3pt,color=txtcolor3] at (2.5mm,0){% 
            \pgfuseplotmark{o}%
        };
        \node (base) at (0,-.5ex) {};
        \end{tikzpicture}%
}
\newcommand{\inpgrad}{%
        \begin{tikzpicture}[inner sep=0pt, baseline=(base)]%
        \draw[txtcolor4,line width=1.5pt](0,0) -- (5mm,0); 
        \node[very thick, mark size=3pt,color=txtcolor4] at (2.5mm,0){% 
            \pgfuseplotmark{square*}%
        };
        \node (base) at (0,-.5ex) {};
        \end{tikzpicture}%
}
\newcommand{\integrad}{%
        \begin{tikzpicture}[inner sep=0pt, baseline=(base)]%
        \draw[txtcolor5,line width=1.5pt](0,0) -- (5mm,0); 
        \node[very thick, mark size=3pt,color=txtcolor5] at (2.5mm,0){% 
            \pgfuseplotmark{asterisk}%
        };
        \node (base) at (0,-.5ex) {};
        \end{tikzpicture}%
}
\newcommand{\rankmask}{%
        \begin{tikzpicture}[inner sep=0pt, baseline=(base)]%
        \draw[txtcolor6,line width=1.5pt](0,0) -- (5mm,0); 
        \node[very thick, mark size=3pt,color=txtcolor6] at (2.5mm,0){% 
            \pgfuseplotmark{*}%
        };
        \node (base) at (0,-.5ex) {};
        \end{tikzpicture}%
}

\newcommand\heightunify{0.88}

\usepackage{microtype}

\aclfinalcopy % Uncomment this line for the final submission
\title{Are Interpretations Fairly Evaluated? \\ A Definition Driven Pipeline for Post-Hoc Interpretability}

\author{Ninghao Liu$^{\dagger}$, Yunsong Meng$^{\ddagger}$, Xia Hu$^{\dagger}$, Tie Wang$^{\ddagger}$, Bo Long$^{\ddagger}$ \\
  $^{\dagger}$Department of Computer Science and Engineering, Texas A\&M University, TX, USA \\
  $^{\ddagger}$LinkedIn, Mountain View, CA, USA \\
  \texttt{\{nhliu43,xiahu\}@tamu.edu, \{yumeng,tiewang,blong\}@linkedin.com} %\\\And
}

\date{}

\begin{document}
\maketitle
\begin{abstract}
Recent years have witnessed an increasing number of interpretation methods being developed for improving transparency of NLP models. Meanwhile, researchers also try to answer the question that whether the obtained interpretation is \textit{faithful} in explaining mechanisms behind model prediction? Specifically, \citep{Jain-Wallace19attNotExplain} proposes that ``attention is not explanation" by comparing attention interpretation with gradient alternatives. However, it raises a new question that can we safely pick one interpretation method as the ground-truth? If not, on what basis can we compare different interpretation methods? In this work, we propose that it is crucial to have a \textit{concrete definition} of interpretation before we could evaluate faithfulness of an interpretation. The definition will affect both the algorithm to obtain interpretation and, more importantly, the metric used in evaluation. Through both theoretical and experimental analysis, we find that although interpretation methods perform differently under a certain evaluation metric, such a difference may not result from interpretation quality or faithfulness, but rather the inherent bias of the evaluation metric.
\end{abstract}
% No matter it is intended to treat the alternatives as ground-truth,
%, while some claims are partially challenged in \citep{Wiegreffe-Pinter19attNotNotExplain} and \citep{Serrano-Smith19isAttIntp} through studies of the existence of adversarial samples

\section{Introduction}
Interpretability is drawing increasing interests for many advanced NLP models. As more complex models achieve state-of-the-art performances and are deployed in real applications, it is crucial to maintain our ability to understand why a particular decision is made by those models. Some commonly used interpretation methods for NLP models include gradient-based methods~\cite{Simonyan-etal13deepInsideCNNsaliency, denil2014extraction, Smilkov-etal18smoothgrad, Wallace-etal19allenNLPInterpret} and attention methods~\cite{Bahdanau-etal15, vaswani2017attention}.

With more interpretation methods at hand, a question naturally arises that: Which method is better? Or to be more precise: Which method is more faithful in explaining the model prediction? A recent work~\cite{Jain-Wallace19attNotExplain} fosters the discussion by discovering a discrepancy between attention-based and gradient-based interpretation, drawing the conclusion that attention modules may not provide meaningful explanations. Also, the authors proposed that, if attention provides a faithful explanation for model predictions, the following two properties should hold: (1) Attention scores should correlate to feature importance measures (e.g., gradients); (2) Counterfactual attention scores ought to yield corresponding changes in model prediction. \citet{Serrano-Smith19isAttIntp} examines the faithfulness of attention through attention weights erasure. \citet{Wiegreffe-Pinter19attNotNotExplain} challenges the second property by proposing a more practical experiment for attention manipulation.

\begin{figure}[t]
\centering
 \includegraphics[scale=0.35]{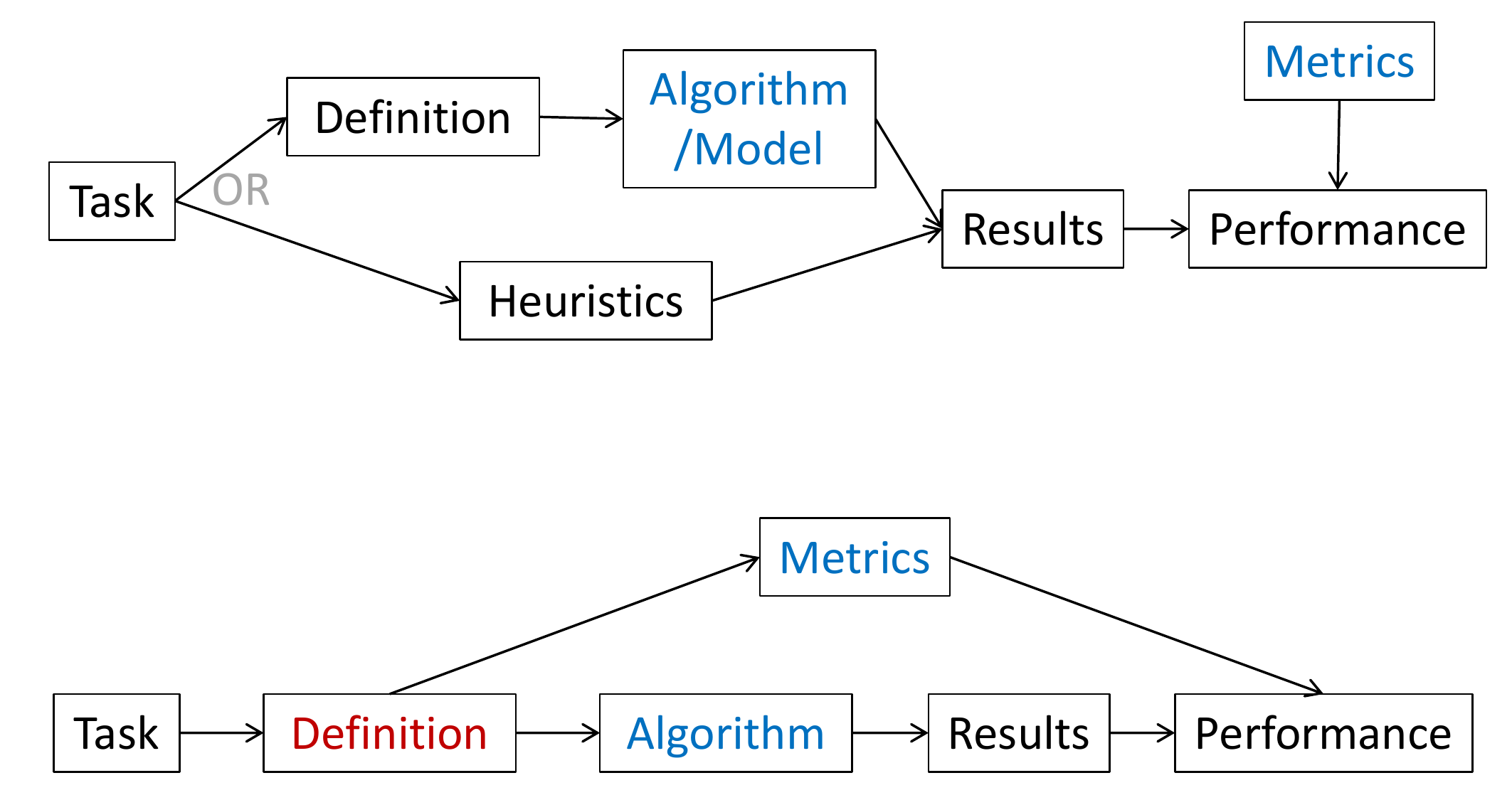}
 \caption{Pipelines of defining, obtaining and evaluating interpretation. Upper: The traditional pipeline. Lower: The proposed DDP pipeline.} \label{fig:pipeline}
 \vspace{-4pt}
\end{figure}

%In a broader sense, we propose that the argument above is related to the definition of interpretation, beyond mere comparison between attention and gradient methods. 
While the discussion above only involves comparing attention with gradient interpretation, it actually relates to a greater challenge, i.e., how to evaluate interpretation. Many existing work follow the pipeline shown in the upper part of Figure~\ref{fig:pipeline}. For example, we may design a new interpretation method based on heuristics~\cite{denil2014extraction, Selvaraju-etal16gradCAM, nourani2019effects}, then compare it with certain baseline methods such as Integrated Gradient~\cite{Sundararajan-etal16integratedGradient} and SmoothGrad~\cite{Smilkov-etal18smoothgrad}, evaluated under a chosen evaluation metric. The main issue for such a pipeline is that, the designed method may not essentially share the same objective
as the metric. Sometimes, even different algorithms in comparison do not share the same objective, such as Integrated Gradient and SmoothGrad, as we will show later in the paper. Another typical issue is choosing a \textit{subjective} metric (related to human cognition habits~\cite{yang2019evaluating}) over a method that targets to extract \textit{objective} and faithful interpretation~\cite{Rudin-18pleasestop}. Such a discrepancy could make algorithms not fairly evaluated, thus leading to a false intuition of algorithm performance. % the algorithm or heuristics could be a specific gradient-based method or its variants~\cite{denil2014extraction, Smilkov-etal18smoothgrad}

%Metaphorically, it is like we are comparing ``the speed of an air plane" with ``the length of a train".
% Therefore, interpretation is simply the result of a specific algorithm that solves the adversarial problem as defined.
In this work, we propose a Definition Driven Pipeline (DDP) which guides the development and evaluation of interpretation algorithms, as shown in the lower part of Figure~\ref{fig:pipeline}. Different from the traditional pipeline, in DDP both the algorithm and the metric are formulated based on the same definition. To be more specific, in this work, we define interpretation of model prediction based on adversarial perturbation~\cite{Good-etal15explaining}, i.e., features are considered as important if their distortions cause significant model prediction changes. We will prove that, by correspondingly choosing constraints in adversary, many existing interpretation algorithms are equivalent to DDP. More importantly, each algorithm corresponds to a metric, which guarantees proper evaluation of interpretation results. Based on DDP, we also discuss how to align interpretation with human-specified rationale. In addition, we show that the validity of the two properties in~\cite{Jain-Wallace19attNotExplain} is closely related to the definition of interpretation. It is worth noting that, although \cite{Wiegreffe-Pinter19attNotNotExplain, jacovi2020towards} provide guidelines on interpretation evaluation, we propose a concrete solution and conduct experiments accordingly.

The contributions of this work are as below:
\begin{itemize}[leftmargin=*, topsep=0pt, noitemsep]
    \item We propose the Definition Driven Pipeline for developing and evaluating interpretation methods for NLP models. We prove that many existing interpretation methods can be derived from DDP.
    \item By extending DDP, we propose a method to align interpretation with human-specified rationale.
    \item Through experiments, we validate the importance of evaluating interpretation using the proper metric, in order to prevent false intuition of interpretation faithfulness. 
\end{itemize}

\section{Definition Driven Pipeline (DDP)}
Formally, an NLP model is represented as a function $f:\mathcal{X}\rightarrow \mathcal{Y}$. The input text is represented as a concatenation of word embeddings, where $\textbf{x}=[\textbf{x}^1; \textbf{x}^2; \dots; \textbf{x}^N]\in \mathcal{X}$ and $\textbf{x}^n$ denotes the embedding of the $n$-th word. For classification tasks, $f(\textbf{x})\in \mathbb{R}^C$, where $C$ is the number of classes and $f_c(\textbf{x})$ is the output probability for the $c$-th class.

Throughout this work, we use \textit{adversarial attack}~\cite{Good-etal15explaining, Fong-Vedaldi17perturbation} to define interpretation. Given the target function $f$, input $\textbf{x}$ and label $c$ of interest, we define the raw interpretation as $h = \textbf{x} - \textbf{x}^*$, where $\textbf{x}^* = [{\textbf{x}^*}^1; {\textbf{x}^*}^2; \dots; {\textbf{x}^*}^N]$. The general solution for obtaining $\textbf{x}^*$ is:
\begin{equation}\label{eq:general}
        \min_{\textbf{x}^*} J(f, \textbf{x}^*, c), \text{\,\,s.t.\,\,} \textbf{x}^*\in domain(\textbf{x}, \mathcal{E}),
\end{equation}
where $J$ is the adversarial objective function, $domain()$ constrains the range of solution, and $\mathcal{E}$ is the set of hyperparameters.
Next, we will show some example definitions and how existing interpretation methods are related to them.

\subsection{Continuous-Space Adversary (CSA)}
The fundamental rationale behind using adversary is that, if the resultant adversarial sample $\textbf{x}^*$ perturbs features to which the model prediction is sensitive, then such a perturbation is expected to weaken the prediction confidence. On the contrary, if insignificant features are perturbed by $\textbf{x}^*$, the prediction is not likely to be affected. The problem is thus defined as: 
\begin{equation}\label{eq:csa}
    \begin{split}
        &\min_{\textbf{x}^*}\, f_c(\textbf{x}^*) - f_c(\textbf{x}) \\
        \text{s.t. }\, &\| \textbf{x}^* - \textbf{x} \|_2 \le \epsilon .
    \end{split}
\end{equation}
where $\epsilon$ is the distance hyperparameter. In CSA, $\textbf{x}^*$ is allowed to locate freely in the continuous neighborhood around the input $\textbf{x}$. The above optimization problem could be solved through various algorithms with different precision levels. Some examples are as below.
\begin{itemize}[leftmargin=*, topsep=0pt, noitemsep]
    \item \textbf{Vanilla Gradient (VaGrad):} The objective above could be optimized by only a single step of gradient descent, followed by the projection into constraint. Thus, $\textbf{x}^* = \textbf{x} - \epsilon \cdot \nabla f_c(\textbf{x})/\|\nabla f_c(\textbf{x})\|_2$, so $h = \textbf{x}-\textbf{x}^* \propto \nabla f_c(\textbf{x})$, which is exactly the Vanilla Gradient explanation~\cite{Simonyan-etal13deepInsideCNNsaliency, hechtlinger2016interpretation}.
    \item \textbf{Smooth Gradient (SmoothGrad):} One drawback of VaGrad is that it suffers from limited precision since function $f_c$ could be noisy. Assume that $f_c$ is subject to Gaussian noise, then the unbiased estimation~\cite{hogg2010probability} of interpretation is $h \propto \sum_{\textbf{x}'} \nabla f_c(\textbf{x}')$, where $\textbf{x}'\sim N(0, \sigma^2)$ is sampled from a Gaussian distribution, thus is the same as SmoothGrad~\cite{Smilkov-etal18smoothgrad}.
    \item \textbf{Iterative Gradient (IterGrad):} Another drawback of VaGrad is there is no guarantee that linear approximation works well around $f_c(\textbf{x})$. Thus, a more precise solution is obtained by iterative optimization:
    \begin{equation}
        \textbf{x}^{(0)} = \textbf{x},\, \textbf{x}^{(t+1)} = Proj(\textbf{x}^{(t)} - \alpha \nabla f_c(\textbf{x}^{(t)})),
    \end{equation}
    where $\textbf{x}^*=\textbf{x}^{(t_{max})}$ and $Proj()$ projects the instance into the $\epsilon$ vicinity of $\textbf{x}$.
\end{itemize}
After obtaining raw interpretation, importance of the $n$-th word is computed as $\| \textbf{x}^n - {\textbf{x}^*}^n \|_2$. Finally, according to the definition in Equation~\ref{eq:csa}, the \textit{metric} for evaluating interpretation faithfulness is naturally set as $f_c(\textbf{x})-f_c(\textbf{x}-h)$ (or $f_c(\textbf{x})-f_c(\textbf{x}^*)$), i.e., the objective function in the definition.

\subsection{Embedding Erasure Adversary (ERA)}
Different from CSA that treats input as a purely continuous-space instance, it is more natural to treat each text as discrete word tokens. In this case, perturbation of input considers each embedding as a unit, and removing a word's contribution can be done by setting its embedding to the zero vector~\cite{Li-Jurafsky17erasure}. Thus, the adversary problem is modified as:
\begin{equation}\label{eq:era}
    \begin{split}
        &\min_{\textbf{x}_{-\mathcal{S}}}\, f_c(\textbf{x}_{-\mathcal{S}}) - f_c(\textbf{x}) \\
        \text{s.t. }\, &| \mathcal{S} | \le s
    \end{split}
\end{equation}
where $\textbf{x}_{-\mathcal{S}}$ is the text after erasing embeddings of words in $\mathcal{S}$, while the number of embeddings erased is limited to $s$. Optimization over discrete space is challenging, and solving such as problem depends on how well we estimate the objective value. Some examples are as below.
\begin{itemize}[leftmargin=*, topsep=0pt, noitemsep]
    \item \textbf{Input Times Gradient (InpGrad):} According to first-order Taylor expansion, $f_c(\textbf{x}_{-\mathcal{S}}) \approx f_c(\textbf{x}) + \nabla f_c(\textbf{x}) \cdot (\textbf{x}_{-\mathcal{S}}-\textbf{x})$, so that the objective is equal to $\nabla f_c(\textbf{x}) \cdot (\textbf{x}_{-\mathcal{S}}-\textbf{x}) = \sum_n \nabla f_c(\textbf{x}^n) \cdot (\textbf{x}^n_{-\mathcal{S}}-\textbf{x}^n)$. Different embeddings contribute independently to the total value. If a word is to be erased, then its contribution equals $-\nabla f_c(\textbf{x}^n) \cdot (\textbf{0}-\textbf{x}^n)=\nabla f_c(\textbf{x}^n)\cdot \textbf{x}^n$, which is exactly the input-times-gradient algorithm~\cite{denil2014extraction}. To select the $s$ most important words, we can use a greedy strategy by ranking words according to their inner product value between gradient and embedding. Finally, the top $s$ words are selected for interpretation.
    \item \textbf{Integrated Gradient (InteGrad):} An obvious limitation for InpGrad is that it is not suitable for functions which cannot be well approximated by first-order Taylor expansion. Another algorithm with better precision is to use piece-wise linear functions for approximation. Let $[\textbf{x}^{(0)}, \textbf{x}^{(1)}, \dots, \textbf{x}^{(T)}]$ be a series of points located along the line between $\textbf{x}_{-\mathcal{S}}$ and $\textbf{x}$, where $\textbf{x}^{(0)}=\textbf{x}$ and $\textbf{x}^{(T)}=\textbf{x}_{-\mathcal{S}}$. The distance between adjacent points is the same. Accordingly, $f_c(\textbf{x}_{-\mathcal{S}}) \approx f_c(\textbf{x}) + \sum^{T-1}_{t=0} \nabla f(\textbf{x}^{(t)}) \cdot (\textbf{x}^{(t+1)}-\textbf{x}^{(t)})$. Thus,
    \begin{equation}
    \begin{split}
        f_c(\textbf{x}_{-\mathcal{S}}) - f_c(\textbf{x}) &\approx \sum^{T-1}_{t=0} \nabla f(\textbf{x}^{(t)}) \cdot (\textbf{x}^{(t+1)}-\textbf{x}^{(t)}) \\
        &= \frac{1}{T}(\textbf{x}_{\mathcal{S}} - \textbf{x}) \cdot \sum^{T-1}_{t=0} \nabla f(\textbf{x}^{(t)}),
    \end{split}
    \end{equation}
    which is essentially the same as InteGrad~\cite{Sundararajan-etal16integratedGradient}. We could first use InteGrad to estimate the contribution of word embeddings, and then select the top $s$ important words as interpretation.
\end{itemize}
Under this definition in Equation~\ref{eq:era}, the \textit{metric} for evaluating interpretation faithfulness should be set as $f_c(\textbf{x}_{-\mathcal{S}}) - f_c(\textbf{x})$. That is, to obtain $\textbf{x}_{-\mathcal{S}}$, we first identify $s$ most important words, change their embeddings to zero vectors, and then compare the output variation. The importance of each individual word $\textbf{x}^n$ is $f_c(\textbf{x}_{-\{n\}})-f_c(\textbf{x})$.

\subsection{Message Masking Adversary (MMA)}
In attention models, attention scores mark the information propagation paths between representations. Different from gradient-based algorithms that directly builds input-output relation, attention scores indicate the \textit{intermediate} relations between latent representations of adjacent layers. 

To craft adversarial samples on attention models, we can assign an external mask entry $m\in \{0,1\}$ to each attention score $a\in \mathbb{R}$ to have a new score $a' = a\cdot m$, where $m=0$ means blocking the message passing. Assume $f_c(\textbf{x})$ could be written as $f^2_c(f^1(\textbf{x}), \textbf{a})$, where $f^1$ maps input into representation, and $f^2_c$ receives representation with attention $\textbf{a}$ for final prediction. The adversary problem is designed as:
\begin{equation}\label{eq:mma}
\begin{split}
    &\min_{\textbf{m}} f^2_c(f^1(\textbf{x}), \textbf{a}\odot \textbf{m}) - f^2_c(f^1(\textbf{x}), \textbf{a}) \\
    \text{s.t. }\, &\| \textbf{1}-\textbf{m} \|_0 \le s, \,\textbf{m}[n]\in \{0,1\}
\end{split}
\end{equation}
where $\textbf{a}$ and $\textbf{m}$ denote attentions and mask vectors, respectively. $\odot$ means element-wise multiplication. $\|\cdot\|_0$ is $L_0$ norm. Each mask entry $\textbf{m}[n]$ is binary. The constraint means no more than $s$ attention scores are blocked.
\begin{itemize}[leftmargin=*, topsep=0pt, noitemsep]
    \item \textbf{Ranked Masking (RankMask):} We first rank the attention scores and choose the highest $s$ scores to be assigned with $m=0$, while the others are given $m=1$.
\end{itemize}
Similar to previous cases, the \textit{metric} under MMA is simply the objective function above. In this work, we use LSTM with attention in Figure~\ref{fig:lstmatt} as the attention model. Word embeddings are seen as important if they are in same positions of top-ranked attentions.

\section{Aligning Interpretation with Human Cognition}\label{sec:align}
In some applications, interpretation accuracy is defined as the matching degree between interpretation and human cognition habits~\cite{Fong-Vedaldi17perturbation, yang2019evaluating}. That is, users may expect models to pay attention to the same set of words as human when making predictions. However, traditional training schemes usually do not consider this requirement. 

In this section, we introduce how to extend DDP to incorporate human cognition into training to improve model interpretability. Specifically, we hope that solving the optimization in Equation~\ref{eq:general} will not perturb the embedding (or attention) of words that are not regarded as important by human. We denote the set of words that are considered as important by human as $\mathcal{I}$. Given an instance $\textbf{x}$ of interest, suppose $\textbf{x}^*$ is the solution after running DDP. Then, another instance is crafted as $\tilde{\textbf{x}}^*$, where 
$\tilde{\textbf{x}}^{*n}=\textbf{x}^n$ if $n\in \mathcal{I}$, and otherwise $\tilde{\textbf{x}}^{*n}={\textbf{x}^*}^n$. The loss for retraining the model is formulated as:
\begin{equation}
    \min_f  \, \sum_{\textbf{x}\in \mathcal{X}} l(\tilde{\textbf{y}}, f(\tilde{\textbf{x}}^*)) ,
\end{equation}
where $\tilde{\textbf{y}}$ is the soft label and $\tilde{\textbf{y}} = f(\textbf{x})$, $l$ is the instance-level cross entropy loss. The idea behind the objective is that we restrict sensitivity of model $f$ (i.e., $f_c$) to those words that are not regarded as important by human. Therefore, in $\tilde{\textbf{x}}^*$, information of important words remains the same as the clean sample $\textbf{x}$, while only unimportant words are perturbed according to ${\textbf{x}^*}^n$.

Such a training scheme is similar to adversarial training~\cite{Good-etal15explaining}, since both try to stabilize predictions before and after input perturbation. Our method is different from traditional adversarial training in two aspects. First, we introduce human cognition knowledge $\mathcal{I}$, which performs post-processing to produce $\tilde{\textbf{x}}^*$ from $\textbf{x}^*$. Second, in language processing, we could make use of the distribution of word embeddings to assist human knowledge. Collecting human knowledge is a laborious task, and usually the result is not fully comprehensive. Thus, we can first collect a seed list of words $\mathcal{I}_{seed}$ provided by human, and then slightly modify the rule of getting $\tilde{\textbf{x}}^*$, where $\tilde{\textbf{x}}^{*n}=\textbf{x}^n$ if $\exists\, i\in \mathcal{I} \land n\in neighbor(i)$. Here $n\in neighbor(i)$ means the embedding of word $n$ is in the neighborhood of the embedding of word $i$. This is because that, although words are discrete symbols, their embeddings are correlated in the continuous latent space.

\begin{table*}[t]
\small
	\centering
      \begin{tabular}[0.1\textwidth]{ccccc}
      \toprule
      \multicolumn{1}{c}{Dataset} & \multicolumn{1}{c}{Avg. Length} & \multicolumn{1}{c}{Train size} & \multicolumn{1}{c}{Valid size} & \multicolumn{1}{c}{Test size} \\
      \multicolumn{1}{c}{} & \multicolumn{1}{c}{} & \multicolumn{1}{c}{(pos/neg)} & \multicolumn{1}{c}{(pos/neg)} & \multicolumn{1}{c}{(pos/neg)} \\
      \hline 
      SST2 & $19$ & $3310$/$3610$  & $428$/$444$ & $912$/$909$  \\
      Yelp & $20$ & $18746$/$31543$  & $2150$/$3537$ & $1432$/$2369$  \\ 
      AGNews & $41$ & $25736$/$25368$  & $2847$/$2831$ & $1823$/$1782$  \\ 
      \bottomrule
      \end{tabular}
     \vspace{-0pt}
	\caption{Datasets statistics.} \label{table:datasets}
\vspace{0pt}
\end{table*}

\section{Experiments}\label{sec:exp1}
\begin{figure}[t]
\centering
 \includegraphics[scale=0.35]{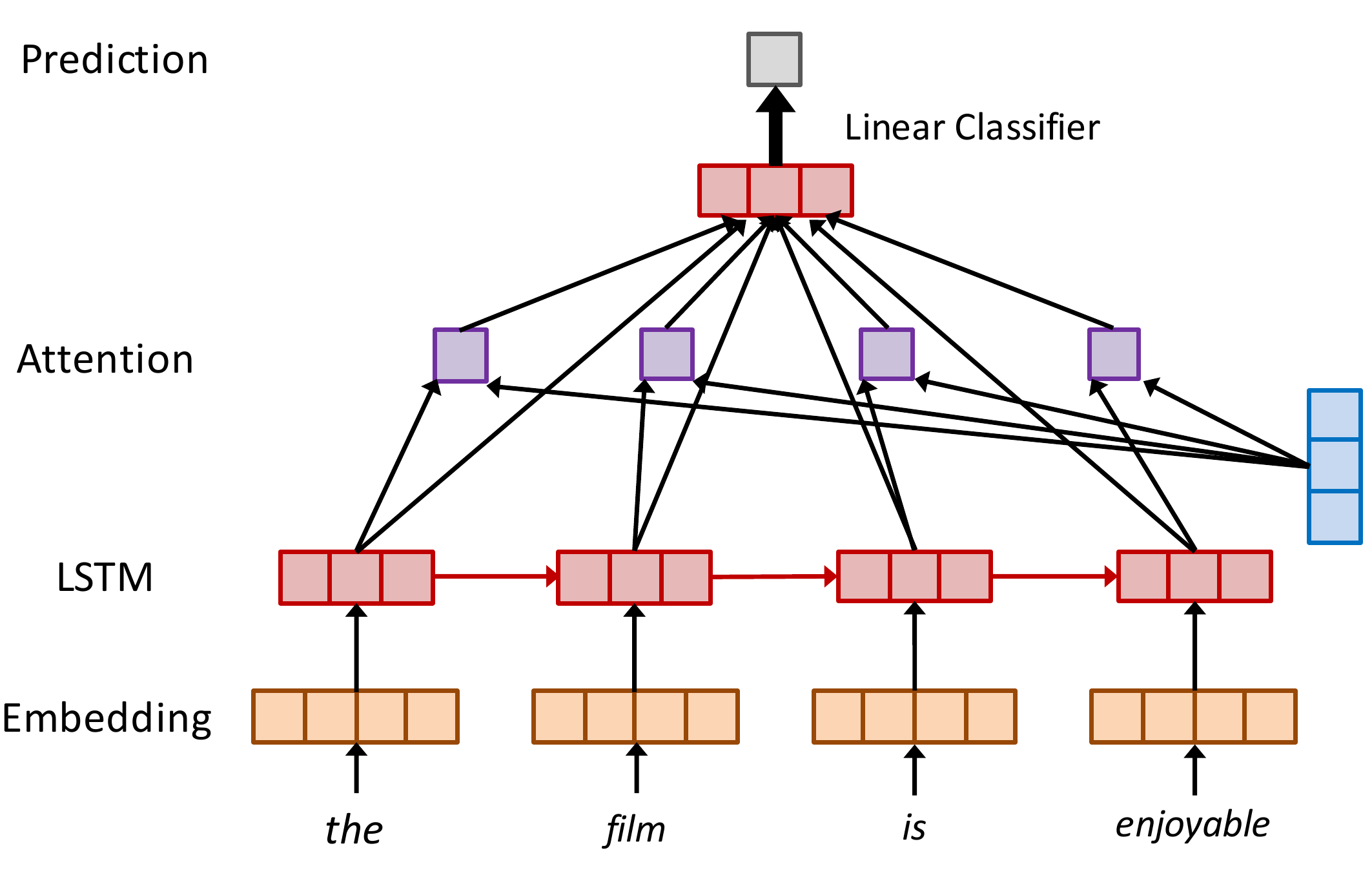}
 \caption{The LSTM\_att model used in experiments.} \label{fig:lstmatt}
\end{figure}

In this section, we compare the ``faithfulness" of interpretation obtained with different definitions, evaluated under different metrics. The evaluation outcome shows that an interpretation method has inborn advantages if its definition matches the metric. It thus suggests that an optimally faithful interpretation may \textit{not} even exist if a concrete definition is not set up in advance.

\begin{table}[t!]
\small
	\centering
      \begin{tabular}[0.1\textwidth]{cccc}
      \toprule
      \multicolumn{1}{c}{Dataset} & \multicolumn{1}{c}{LSTM} & \multicolumn{1}{c}{LSTM\_att} & \multicolumn{1}{c}{BERT}  \\
      \hline 
      SST2 & $0.802$ & $0.812$  & $0.907$  \\
      Yelp & $0.931$  & $0.929$  & $0.962$  \\ 
      AGNews & $0.926$  & $0.923$  & $0.952$  \\ 
      \bottomrule
      \end{tabular}
     \vspace{-0pt}
	\caption{Test performance on trained models.} \label{table:test}
\vspace{-4pt}
\end{table}

% LSTM: CSA, ERA
% Bert: CSA, ERA
% LSTM_att: CSA, ERA, MMA
\begin{figure*}[t]
        \centering
        \begin{subfigure}[b]{0.23\textwidth}
        \setlength\figureheight{\heightunify in}
        \setlength\figurewidth{1.15in}
        \centering  \scriptsize
        % This file was created by matlab2tikz.
%
%The latest updates can be retrieved from
%  http://www.mathworks.com/matlabcentral/fileexchange/22022-matlab2tikz-matlab2tikz
%where you can also make suggestions and rate matlab2tikz.
%
\definecolor{mycolor1}{rgb}{1,0.65,0}%
\definecolor{mycolor2}{rgb}{1,0.94,0}%
\definecolor{mycolor3}{rgb}{1.0, 0.25, 0.25}%
\definecolor{mycolor4}{rgb}{0.39, 0.58, 0.93}
\definecolor{mycolor5}{rgb}{0., 0.18, 0.39}

\begin{tikzpicture}

\begin{axis}[%
width=0.951\figurewidth,
height=\figureheight,
at={(0\figurewidth,0\figureheight)},
scale only axis,
scaled x ticks=true,
xticklabels={0.25, 0.50, 0.75, 1.00},
xtick={1,2,3,4},
xmin=1,
xmax=4,
xlabel style={font=\color{white!15!black}},
xlabel={$\epsilon$},
ymin=0.00,
ymax=0.60,
grid, % --added
grid style={line width=.15pt, draw=gray!15}, % --added, dashed, 
ylabel style={font=\color{white!15!black}},
ylabel={CSA metric},
axis background/.style={fill=white},
axis background/.style={fill=white} % --added
]
\addplot [color=mycolor1, mark=asterisk, mark options={solid, mycolor1}, thick]
  table[row sep=crcr]{%
1	0.119537\\
2	0.248420\\
3	0.361565\\
4	0.454075\\
};
%\addlegendentry{RAND}

\addplot [color=mycolor2, mark=triangle, mark options={solid, mycolor2}, thick]
  table[row sep=crcr]{%
1	0.119543\\
2	0.248470\\
3	0.361691\\
4	0.454316\\
};
%0.12109, 0.25788, 0.37746, 0.47231
%\addlegendentry{SEP}

\addplot [color=mycolor3, mark=o, mark options={solid, mycolor3}, thick]
  table[row sep=crcr]{%
1	0.126231\\
2	0.268208\\
3	0.392697\\
4	0.511667\\
};
%0.12813, 0.27969, 0.41367, 0.53209
%\addlegendentry{DFE}

\addplot [color=mycolor4, mark=square*, mark options={solid, mycolor4}, thick]
  table[row sep=crcr]{%
1	0.010300\\
2	0.028571\\
3	0.048755\\
4	0.068073\\
};
%0.04323, 0.08623, 0.12550, 0.15896
%\addlegendentry{DFE}

\addplot [color=mycolor5, mark=asterisk, mark options={solid, mycolor5}, thick]
  table[row sep=crcr]{%
1	0.012837\\
2	0.034242\\
3	0.063552\\
4	0.097787\\
};

\end{axis}
\end{tikzpicture}% 
            \caption[Network2]%
            {{\small LSTM / CSA}}
        \end{subfigure}
        \hfill
        \begin{subfigure}[b]{0.23\textwidth}
        \setlength\figureheight{\heightunify in}
        \setlength\figurewidth{1.15in}
        \centering  \scriptsize
        % This file was created by matlab2tikz.
%
%The latest updates can be retrieved from
%  http://www.mathworks.com/matlabcentral/fileexchange/22022-matlab2tikz-matlab2tikz
%where you can also make suggestions and rate matlab2tikz.
%
\definecolor{mycolor1}{rgb}{1,0.65,0}%
\definecolor{mycolor2}{rgb}{1,0.94,0}%
\definecolor{mycolor3}{rgb}{1.0, 0.25, 0.25}%
\definecolor{mycolor4}{rgb}{0.39, 0.58, 0.93}
\definecolor{mycolor5}{rgb}{0., 0.18, 0.39}

\begin{tikzpicture}

\begin{axis}[%
width=0.951\figurewidth,
height=\figureheight,
at={(0\figurewidth,0\figureheight)},
scale only axis,
scaled x ticks=true,
xticklabels={1,2,3,4},
xtick={1,2,3,4},
xmin=1,
xmax=4,
xlabel style={font=\color{white!15!black}},
xlabel={$s$},
ymin=0.00,
ymax=0.60,
grid, % --added
grid style={line width=.15pt, draw=gray!15}, % --added, dashed, 
ylabel style={font=\color{white!15!black}},
ylabel={ERA metric},
axis background/.style={fill=white},
axis background/.style={fill=white} % --added
]
\addplot [color=mycolor1, mark=asterisk, mark options={solid, mycolor1}, thick]
  table[row sep=crcr]{%
1	0.03948179\\
2	0.07605552\\
3	0.12601113\\
4	0.15625506\\
};
%0.10019287, 0.13370584, 0.1685176, 0.19753554
%\addlegendentry{VaGrad}

\addplot [color=mycolor2, mark=triangle, mark options={solid, mycolor2}, thick]
  table[row sep=crcr]{%
1	0.03899756\\
2	0.07818375\\
3	0.12621341\\
4	0.16029641\\
};
%0.10019287, 0.13358119, 0.16842477, 0.19651913
% 0.03899756 0.07818375 0.12621341 0.16029641
%\addlegendentry{SmoothGrad}

\addplot [color=mycolor3, mark=o, mark options={solid, mycolor3}, thick]
  table[row sep=crcr]{%
1	0.04568086\\
2	0.09267365\\
3	0.14821541\\
4	0.18729965\\
};
%0.10524706, 0.13824082, 0.1767163, 0.21206722
%0.04568086 0.09267365 0.14821541 0.18729965
%\addlegendentry{IterGrad}

\addplot [color=mycolor4, mark=square*, mark options={solid, mycolor4}, thick]
  table[row sep=crcr]{%
1	0.17992776\\
2	0.32348302\\
3	0.40808931\\
4	0.46268828\\
};
%0.17992776 0.32348302 0.40808931 0.46268828
%0.21993248 0.3877712  0.48477083 0.55185667
%\addlegendentry{InpGrad}

\addplot [color=mycolor5, mark=asterisk, mark options={solid, mycolor5}, thick]
  table[row sep=crcr]{%
1	0.19807485\\
2	0.35524494\\
3	0.46730655\\
4	0.54360266\\
};
%0.19807485 0.35524494 0.46730655 0.54360266
%\addlegendentry{InteGrad}

\end{axis}
\end{tikzpicture}% 
            \caption[Network2]%
            {{\small LSTM / ERA}}
        \end{subfigure}
        \hfill
        %\vskip\baselineskip
        \begin{subfigure}[b]{0.23\textwidth}  
            \setlength\figureheight{\heightunify in}
        \setlength\figurewidth{1.15in}
        \centering  \scriptsize
        % This file was created by matlab2tikz.
%
%The latest updates can be retrieved from
%  http://www.mathworks.com/matlabcentral/fileexchange/22022-matlab2tikz-matlab2tikz
%where you can also make suggestions and rate matlab2tikz.
%
\definecolor{mycolor1}{rgb}{1,0.65,0}%
\definecolor{mycolor2}{rgb}{1,0.94,0}%
\definecolor{mycolor3}{rgb}{1.0, 0.25, 0.25}%
\definecolor{mycolor4}{rgb}{0.39, 0.58, 0.93}
\definecolor{mycolor5}{rgb}{0., 0.18, 0.39}

\begin{tikzpicture}

\begin{axis}[%
width=0.951\figurewidth,
height=\figureheight,
at={(0\figurewidth,0\figureheight)},
scale only axis,
scaled x ticks=true,
xticklabels={0.05, 0.10, 0.15, 0.20},
xtick={1,2,3,4},
xmin=1,
xmax=4,
xlabel style={font=\color{white!15!black}},
xlabel={$\epsilon$},
ymin=0.00,
ymax=0.60,
grid, % --added
grid style={line width=.15pt, draw=gray!15}, % --added, dashed, 
ylabel style={font=\color{white!15!black}},
ylabel={CSA metric},
axis background/.style={fill=white},
axis background/.style={fill=white} % --added
]
\addplot [color=mycolor1, mark=asterisk, mark options={solid, mycolor1}, thick]
  table[row sep=crcr]{%
1	0.09258\\
2	0.20991\\
3	0.30658\\
4	0.37879\\
};
%0.09258, 0.20991, 0.30658, 0.37879
%\addlegendentry{RAND}

\addplot [color=mycolor2, mark=triangle, mark options={solid, mycolor2}, thick]
  table[row sep=crcr]{%
1	0.09259\\
2	0.20992\\
3	0.30663\\
4	0.37881\\
};
%0.09259, 0.20992, 0.30663, 0.37881
%\addlegendentry{SEP}

\addplot [color=mycolor3, mark=o, mark options={solid, mycolor3}, thick]
  table[row sep=crcr]{%
1	0.12411\\
2	0.26307\\
3	0.41530\\
4	0.53245\\
};
%0.12813, 0.27969, 0.41367, 0.53209
%\addlegendentry{DFE}

\addplot [color=mycolor4, mark=square*, mark options={solid, mycolor4}, thick]
  table[row sep=crcr]{%
1	0.01001\\
2	0.03228\\
3	0.05354\\
4	0.072007\\
};
%0.01001, 0.03228, 0.05354, 0.072007
%\addlegendentry{DFE}

\addplot [color=mycolor5, mark=asterisk, mark options={solid, mycolor5}, thick]
  table[row sep=crcr]{%
1	0.00744\\
2	0.02771\\
3	0.046487\\
4	0.059286\\
};
% 0.00744, 0.02771, 0.046487, 0.059286

\end{axis}
\end{tikzpicture}% 
            \caption[]%
            {{\small BERT / CSA}}
        \end{subfigure}
        \hfill
        \begin{subfigure}[b]{0.23\textwidth}  
            \setlength\figureheight{\heightunify in}
        \setlength\figurewidth{1.15in}
        \centering  \scriptsize
        % This file was created by matlab2tikz.
%
%The latest updates can be retrieved from
%  http://www.mathworks.com/matlabcentral/fileexchange/22022-matlab2tikz-matlab2tikz
%where you can also make suggestions and rate matlab2tikz.
%
\definecolor{mycolor1}{rgb}{1,0.65,0}%
\definecolor{mycolor2}{rgb}{1,0.94,0}%
\definecolor{mycolor3}{rgb}{1.0, 0.25, 0.25}%
\definecolor{mycolor4}{rgb}{0.39, 0.58, 0.93}
\definecolor{mycolor5}{rgb}{0., 0.18, 0.39}

\begin{tikzpicture}

\begin{axis}[%
width=0.951\figurewidth,
height=\figureheight,
at={(0\figurewidth,0\figureheight)},
scale only axis,
scaled x ticks=true,
xticklabels={1,2,3,4},
xtick={1,2,3,4},
xmin=1,
xmax=4,
xlabel style={font=\color{white!15!black}},
xlabel={$s$},
ymin=0.00,
ymax=0.60,
grid, % --added
grid style={line width=.15pt, draw=gray!15}, % --added, dashed, 
ylabel style={font=\color{white!15!black}},
ylabel={ERA metric},
axis background/.style={fill=white},
axis background/.style={fill=white} % --added
]
\addplot [color=mycolor1, mark=asterisk, mark options={solid, mycolor1}, thick]
  table[row sep=crcr]{%
1	0.12057786\\
2	0.18185633\\
3	0.22508102\\
4	0.27534258\\
};
%0.12057786, 0.18185633, 0.22508102, 0.27534258
%\addlegendentry{RAND}

\addplot [color=mycolor2, mark=triangle, mark options={solid, mycolor2}, thick]
  table[row sep=crcr]{%
1	0.12057786\\
2	0.18185633\\
3	0.22508102\\
4	0.27535298\\
};
%0.12057786, 0.18185633, 0.22508102, 0.27535298
%\addlegendentry{SEP}

\addplot [color=mycolor3, mark=o, mark options={solid, mycolor3}, thick]
  table[row sep=crcr]{%
1	0.12409556\\
2	0.18188833\\
3	0.22819398\\
4	0.27837211\\
};
%0.12409556, 0.18188833, 0.22819398, 0.27837211
%\addlegendentry{DFE}

\addplot [color=mycolor4, mark=square*, mark options={solid, mycolor4}, thick]
  table[row sep=crcr]{%
1	0.08384714\\
2	0.11675363\\
3	0.14601074\\
4	0.18890988\\
};
%0.08384714, 0.11675363, 0.14601074, 0.18890988
%\addlegendentry{DFE}

\addplot [color=mycolor5, mark=asterisk, mark options={solid, mycolor5}, thick]
  table[row sep=crcr]{%
1	0.17686895\\
2	0.32740351\\
3	0.39470266\\
4	0.45144566\\
};
%0.17686895, 0.32740351, 0.39470266, 0.45144566

\end{axis}
\end{tikzpicture}% 
            \caption[]%
            {{\small BERT / ERA}}
        \end{subfigure}

\vskip\baselineskip

        \hfill
        \begin{subfigure}[b]{0.25\textwidth}
        \setlength\figureheight{\heightunify in}
        \setlength\figurewidth{1.15in}
        \centering  \scriptsize
        % This file was created by matlab2tikz.
%
%The latest updates can be retrieved from
%  http://www.mathworks.com/matlabcentral/fileexchange/22022-matlab2tikz-matlab2tikz
%where you can also make suggestions and rate matlab2tikz.
%
\definecolor{mycolor1}{rgb}{1,0.65,0}%
\definecolor{mycolor2}{rgb}{1,0.94,0}%
\definecolor{mycolor3}{rgb}{1.0, 0.25, 0.25}%
\definecolor{mycolor4}{rgb}{0.39, 0.58, 0.93}
\definecolor{mycolor5}{rgb}{0., 0.18, 0.39}

\begin{tikzpicture}

\begin{axis}[%
width=0.951\figurewidth,
height=\figureheight,
at={(0\figurewidth,0\figureheight)},
scale only axis,
scaled x ticks=true,
xticklabels={0.25, 0.50, 0.75, 1.00},
xtick={1,2,3,4},
xmin=1,
xmax=4,
xlabel style={font=\color{white!15!black}},
xlabel={$\epsilon$},
ymin=0.00,
ymax=0.75,
grid, % --added
grid style={line width=.15pt, draw=gray!15}, % --added, dashed, 
ylabel style={font=\color{white!15!black}},
ylabel={CSA metric},
axis background/.style={fill=white},
axis background/.style={fill=white} % --added
]
\addplot [color=mycolor1, mark=asterisk, mark options={solid, mycolor1}, thick]
  table[row sep=crcr]{%
1	0.184991\\
2	0.378417\\
3	0.523344\\
4	0.625889\\
};
%\addlegendentry{RAND}

\addplot [color=mycolor2, mark=triangle, mark options={solid, mycolor2}, thick]
  table[row sep=crcr]{%
1	0.185025\\
2	0.37858\\
3	0.523579\\
4	0.626052\\
};
%0.12109, 0.25788, 0.37746, 0.47231
%\addlegendentry{SEP}

\addplot [color=mycolor3, mark=o, mark options={solid, mycolor3}, thick]
  table[row sep=crcr]{%
1	0.197726\\
2	0.420919\\
3	0.593982\\
4	0.712531\\
};
%0.12813, 0.27969, 0.41367, 0.53209
%\addlegendentry{DFE}

\addplot [color=mycolor4, mark=square*, mark options={solid, mycolor4}, thick]
  table[row sep=crcr]{%
1	0.022594\\
2	0.062474\\
3	0.108438\\
4	0.152881\\
};
%0.04323, 0.08623, 0.12550, 0.15896
%\addlegendentry{DFE}

\addplot [color=mycolor5, mark=asterisk, mark options={solid, mycolor5}, thick]
  table[row sep=crcr]{%
1	0.022771\\
2	0.062763\\
3	0.117205\\
4	0.177764\\
};

\end{axis}
\end{tikzpicture}% 
            \caption[Network2]%
            {{\small LSTM\_att / CSA}}
        \end{subfigure}
        \hfill
        \begin{subfigure}[b]{0.25\textwidth}  
            \setlength\figureheight{\heightunify in}
        \setlength\figurewidth{1.15in}
        \centering  \scriptsize
        % This file was created by matlab2tikz.
%
%The latest updates can be retrieved from
%  http://www.mathworks.com/matlabcentral/fileexchange/22022-matlab2tikz-matlab2tikz
%where you can also make suggestions and rate matlab2tikz.
%
\definecolor{mycolor1}{rgb}{1,0.65,0}%
\definecolor{mycolor2}{rgb}{1,0.94,0}%
\definecolor{mycolor3}{rgb}{1.0, 0.25, 0.25}%
\definecolor{mycolor4}{rgb}{0.39, 0.58, 0.93}
\definecolor{mycolor5}{rgb}{0., 0.18, 0.39}
\definecolor{mycolor6}{rgb}{0.01, 0.75, 0.24}

\begin{tikzpicture}

\begin{axis}[%
width=0.951\figurewidth,
height=\figureheight,
at={(0\figurewidth,0\figureheight)},
scale only axis,
scaled x ticks=true,
xticklabels={1,2,3,4},
xtick={1,2,3,4},
xmin=1,
xmax=4,
xlabel style={font=\color{white!15!black}},
xlabel={$s$},
ymin=0.00,
ymax=0.70,
grid, % --added
grid style={line width=.15pt, draw=gray!15}, % --added, dashed, 
ylabel style={font=\color{white!15!black}},
ylabel={ERA metric},
axis background/.style={fill=white},
axis background/.style={fill=white} % --added
]
\addplot [color=mycolor1, mark=asterisk, mark options={solid, mycolor1}, thick]
  table[row sep=crcr]{%
1	0.09015038\\
2	0.1455543\\
3	0.17482103\\
4	0.20673918\\
};
%0.09015038 0.1455543  0.17482103 0.20673918
%\addlegendentry{RAND}

\addplot [color=mycolor2, mark=triangle, mark options={solid, mycolor2}, thick]
  table[row sep=crcr]{%
1	0.08904017\\
2	0.14580222\\
3	0.17217874\\
4	0.20864453\\
};
%0.08904017 0.14580222 0.17217874 0.20864453
%\addlegendentry{SEP}

\addplot [color=mycolor3, mark=o, mark options={solid, mycolor3}, thick]
  table[row sep=crcr]{%
1	0.07398941\\
2	0.1270859\\
3	0.16556027\\
4	0.19996566\\
};
%0.07398941 0.1270859  0.16556027 0.19996566
%\addlegendentry{DFE}

\addplot [color=mycolor4, mark=square*, mark options={solid, mycolor4}, thick]
  table[row sep=crcr]{%
1	0.23334137\\
2	0.39715452\\
3	0.5098031\\
4	0.57981006\\
};
%0.23334137 0.39715452 0.5098031  0.57981006
%\addlegendentry{DFE}

\addplot [color=mycolor5, mark=asterisk, mark options={solid, mycolor5}, thick]
  table[row sep=crcr]{%
1	0.24073442\\
2	0.41962398\\
3	0.53938777\\
4	0.61832982\\
};
%0.24073442 0.41962398 0.53938777 0.61832982

\addplot [color=mycolor6, mark=*, mark options={solid, mycolor6}, thick]
  table[row sep=crcr]{%
1	0.10282827\\
2	0.17033383\\
3	0.22782632\\
4	0.25889525\\
};
%0.10282827 0.17033383 0.22782632 0.25889525

\end{axis}
\end{tikzpicture}% 
            \caption[]%
            {{\small LSTM\_att / ERA}}
        \end{subfigure}
        \hfill
        \begin{subfigure}[b]{0.25\textwidth}  
            \setlength\figureheight{\heightunify in}
        \setlength\figurewidth{1.15in}
        \centering  \scriptsize
        % This file was created by matlab2tikz.
%
%The latest updates can be retrieved from
%  http://www.mathworks.com/matlabcentral/fileexchange/22022-matlab2tikz-matlab2tikz
%where you can also make suggestions and rate matlab2tikz.
%
\definecolor{mycolor1}{rgb}{1,0.65,0}%
\definecolor{mycolor2}{rgb}{1,0.94,0}%
\definecolor{mycolor3}{rgb}{1.0, 0.25, 0.25}%
\definecolor{mycolor4}{rgb}{0.39, 0.58, 0.93}
\definecolor{mycolor5}{rgb}{0., 0.18, 0.39}
\definecolor{mycolor6}{rgb}{0.01, 0.75, 0.24}

\begin{tikzpicture}

\begin{axis}[%
width=0.951\figurewidth,
height=\figureheight,
at={(0\figurewidth,0\figureheight)},
scale only axis,
scaled x ticks=true,
xticklabels={1,2,3,4},
xtick={1,2,3,4},
xmin=1,
xmax=4,
xlabel style={font=\color{white!15!black}},
xlabel={$s$},
ymin=0.0,
ymax=0.35,
grid, % --added
grid style={line width=.15pt, draw=gray!15}, % --added, dashed, 
ylabel style={font=\color{white!15!black}},
ylabel={MMA metric},
axis background/.style={fill=white},
axis background/.style={fill=white} % --added
]
\addplot [color=mycolor1, mark=asterisk, mark options={solid, mycolor1}, thick]
  table[row sep=crcr]{%
1	0.09989896\\
2	0.15335714\\
3	0.19084221\\
4	0.21332609\\
};
%0.09989896 0.15335714 0.19084221 0.21332609
%\addlegendentry{RAND}

\addplot [color=mycolor2, mark=triangle, mark options={solid, mycolor2}, thick]
  table[row sep=crcr]{%
1	0.09936597\\
2	0.15327134\\
3	0.18795939\\
4	0.2138888\\
};
%0.09936597 0.15327134 0.18795939 0.2138888
%\addlegendentry{SEP}

\addplot [color=mycolor3, mark=o, mark options={solid, mycolor3}, thick]
  table[row sep=crcr]{%
1	0.07766633\\
2	0.13275706\\
3	0.17443197\\
4	0.20992358\\
};
%0.07766633 0.13275706 0.17443197 0.20992358
%\addlegendentry{DFE}

\addplot [color=mycolor4, mark=square*, mark options={solid, mycolor4}, thick]
  table[row sep=crcr]{%
1	0.11263047\\
2	0.165877\\
3	0.19960532\\
4	0.22795982\\
};
%0.11263047 0.165877   0.19960532 0.22795982
%\addlegendentry{DFE}

\addplot [color=mycolor5, mark=asterisk, mark options={solid, mycolor5}, thick]
  table[row sep=crcr]{%
1	0.10381098\\
2	0.15426908\\
3	0.19264116\\
4	0.2258862\\
};
%0.10381098 0.15426908 0.19264116 0.2258862

\addplot [color=mycolor6, mark=*, mark options={solid, mycolor6}, thick]
  table[row sep=crcr]{%
1	0.12292231\\
2	0.20690539\\
3	0.25852815\\
4	0.29530643\\
};
%0.12292231 0.20690539 0.25852815 0.29530643

\end{axis}
\end{tikzpicture}% 
            \caption[]%
            {{\small LSTM\_att / MMA}}
        \end{subfigure}
        \hfill
        \caption[ The average and standard deviation of critical parameters ]
        {Interpretation faithfulness comparison under different metrics on SST2 dataset. \vagrad:VaGrad,\, \smoothgrad:SmoothGrad,\, \itergrad:IterGrad,\, \inpgrad:InpGrad,\, \integrad:InteGrad,\, \rankmask:RankMask.} 
        \label{fig:sst2_cross}
\end{figure*}
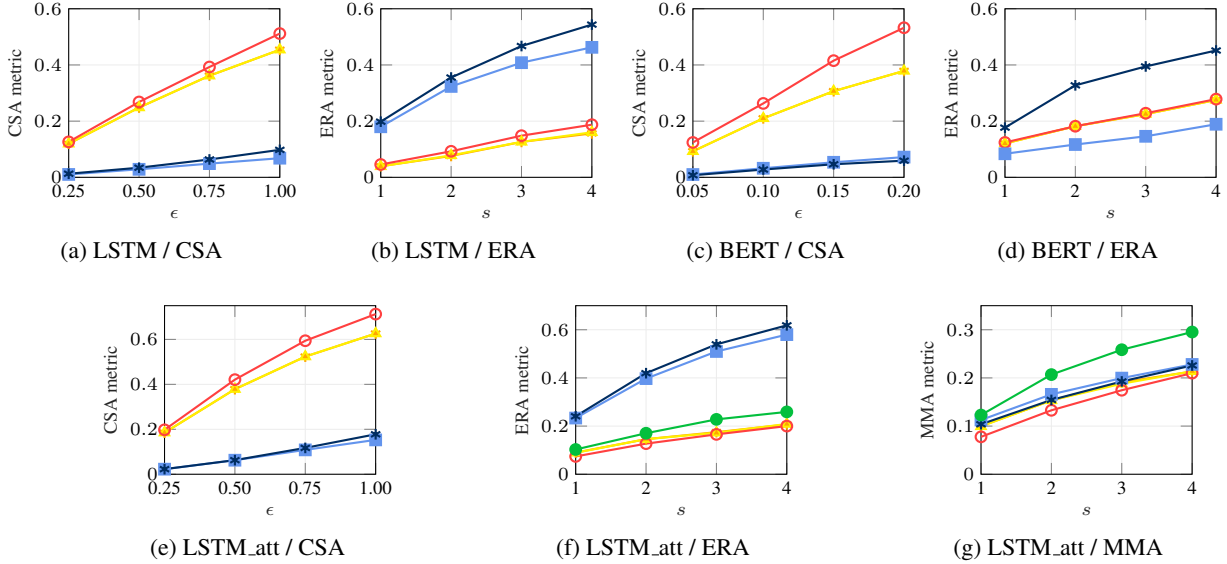

\subsection{Experimental Setup}
In this part, we set up our experiments with with binary classification tasks, and on models with LSTM~\cite{hochreiter1997long}, BERT~\cite{devlin2018bert} and LSTM\_att (Figure~\ref{fig:lstmatt}). This follows previous work~\cite{Wiegreffe-Pinter19attNotNotExplain, Serrano-Smith19isAttIntp} where classification is set as the major task scenario. Future work may extend experiments to more tasks such as question answering or natural language inference.

We conduct experiments on datasets as follows: Stanford Sentiment Treebank2 (SST2)~\cite{socher2013recursive}, Yelp Polarity (Yelp)~\cite{zhang2015character}, and AG NEWS Corpus (AGNews)~\cite{Jain-Wallace19attNotExplain}. The task is to predict sentiment from sentences as positive or negative in SST2 and Yelp, and to predict topic from articles as world (neg.) or business (pos.) in AGNews. We split each dataset into training, validation and testing data. All datasets are in English. Data statistics are listed in Table~\ref{table:datasets}. Since we will perform embedding erasure, texts of length smaller than $5$ have been discarded. Also, although Yelp dataset is not very balanced, it still has similar evaluation results with other datasets. The classification performances on test data for each dataset are listed in Table~\ref{table:test}.

\subsection{Interpretation Faithfulness Comparison of Different Definitions and Algorithms}\label{sec:exp1_result}
Given the target trained model $f$, we obtain interpretation for a set of instances sampled from testing data, regarding why each test instance $\textbf{x}$ is classified with label $c^* = \argmax_c f_c(\textbf{x})$. The interpretation methods include: (1) VaGrad, SmoothGrad and IterGrad defined under CSA; (2) InpGrad and InteGrad defined under ERA; (3) RankMask defined under MMA. After obtaining interpretation from all the method, three rounds of evaluation are conducted using the metric based on CSA, ERA and MMA, respectively.

The default settings of some interpretation methods are as follows. For SmoothGrad, it averages $20$ gradients around each input. For IterGrad, the perturbation step of each iteration is set as $\epsilon/25$. For InteGrad, we place $19$ points uniformly along the path between input and zero-embedding baseline, i.e., $T=20$.

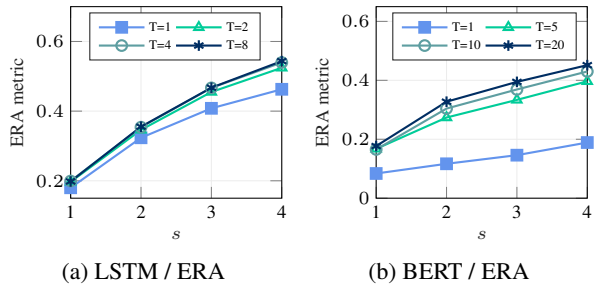
\begin{figure}[t]
        \centering
        \begin{subfigure}[b]{0.23\textwidth}
        \setlength\figureheight{1.00in}
        \setlength\figurewidth{1.15in}
        \centering  \scriptsize
        % This file was created by matlab2tikz.
%
%The latest updates can be retrieved from
%  http://www.mathworks.com/matlabcentral/fileexchange/22022-matlab2tikz-matlab2tikz
%where you can also make suggestions and rate matlab2tikz.
%
\definecolor{mycolor1}{rgb}{1,0.65,0}%
\definecolor{mycolor2}{rgb}{0.39, 0.58, 0.93}%
\definecolor{mycolor3}{rgb}{0.0, 0.8, 0.6}%
\definecolor{mycolor4}{rgb}{0.37, 0.62, 0.63}
\definecolor{mycolor5}{rgb}{0., 0.18, 0.39}

\begin{tikzpicture}

\begin{axis}[%
width=0.951\figurewidth,
height=\figureheight,
at={(0\figurewidth,0\figureheight)},
scale only axis,
scaled x ticks=true,
xticklabels={1,2,3,4},
xtick={1,2,3,4},
xmin=1,
xmax=4,
xlabel style={font=\color{white!15!black}},
xlabel={$s$},
ymin=0.15,
ymax=0.70,
grid, % --added
grid style={line width=.15pt, draw=gray!15}, % --added, dashed, 
ylabel style={font=\color{white!15!black}},
ylabel={ERA metric},
axis background/.style={fill=white},
legend columns = 2,
legend style={legend cell align=left, align=left, draw=white!15!black, nodes={scale=0.7}, at={(0.90, 0.97)}},
axis background/.style={fill=white} % --added
]
\addplot [color=mycolor2, mark=square*, mark options={solid, mycolor2}, thick]
  table[row sep=crcr]{%
1	0.17992776\\
2	0.32348302\\
3	0.40808931\\
4	0.46268828\\
};
%0.17992776 0.32348302 0.40808931 0.46268828
\addlegendentry{T=1}

\addplot [color=mycolor3, mark=triangle, mark options={solid, mycolor3}, thick]
  table[row sep=crcr]{%
1	0.1969755\\
2	0.34623118\\
3	0.45489708\\
4	0.52438925\\
};
%0.1969755  0.34623118 0.45489708 0.52438925
\addlegendentry{T=2}

\addplot [color=mycolor4, mark=o, mark options={solid, mycolor4}, thick]
  table[row sep=crcr]{%
1	0.19930865\\
2	0.35388363\\
3	0.46674313\\
4	0.53965465\\
};
%0.19930865 0.35388363 0.46674313 0.53965465
\addlegendentry{T=4}

\addplot [color=mycolor5, mark=asterisk, mark options={solid, mycolor5}, thick]
  table[row sep=crcr]{%
1	0.19807485\\
2	0.35524494\\
3	0.46730655\\
4	0.54360266\\
};
%0.19807485 0.35524494 0.46730655 0.54360266
\addlegendentry{T=8}

\end{axis}
\end{tikzpicture}% 
            \caption[Network2]%
            {{\small LSTM / ERA}}
        \end{subfigure}
        \hfill
        \begin{subfigure}[b]{0.23\textwidth}
        \setlength\figureheight{1.00in}
        \setlength\figurewidth{1.15in}
        \centering  \scriptsize
        % This file was created by matlab2tikz.
%
%The latest updates can be retrieved from
%  http://www.mathworks.com/matlabcentral/fileexchange/22022-matlab2tikz-matlab2tikz
%where you can also make suggestions and rate matlab2tikz.
%
\definecolor{mycolor1}{rgb}{1,0.65,0}%
\definecolor{mycolor2}{rgb}{0.39, 0.58, 0.93}%
\definecolor{mycolor3}{rgb}{0.0, 0.8, 0.6}%
\definecolor{mycolor4}{rgb}{0.37, 0.62, 0.63}
\definecolor{mycolor5}{rgb}{0., 0.18, 0.39}

\begin{tikzpicture}

\begin{axis}[%
width=0.951\figurewidth,
height=\figureheight,
at={(0\figurewidth,0\figureheight)},
scale only axis,
scaled x ticks=true,
xticklabels={1,2,3,4},
xtick={1,2,3,4},
xmin=1,
xmax=4,
xlabel style={font=\color{white!15!black}},
xlabel={$s$},
ymin=0.00,
ymax=0.65,
grid, % --added
grid style={line width=.15pt, draw=gray!15}, % --added, dashed, 
ylabel style={font=\color{white!15!black}},
ylabel={ERA metric},
axis background/.style={fill=white},
legend columns = 2,
legend style={legend cell align=left, align=left, draw=white!15!black, nodes={scale=0.7}, at={(0.95, 0.97)}},
axis background/.style={fill=white} % --added
]
\addplot [color=mycolor2, mark=square*, mark options={solid, mycolor2}, thick]
  table[row sep=crcr]{%
1	0.08384714\\
2	0.11675363\\
3	0.14601074\\
4	0.18890988\\
};
%0.08384714 0.11675363 0.14601074 0.18890988
\addlegendentry{T=1}

\addplot [color=mycolor3, mark=triangle, mark options={solid, mycolor3}, thick]
  table[row sep=crcr]{%
1	0.16746232\\
2	0.27369746\\
3	0.33361747\\
4	0.39662887\\
};
%0.16746232 0.27369746 0.33361747 0.39662887
\addlegendentry{T=5}

\addplot [color=mycolor4, mark=o, mark options={solid, mycolor4}, thick]
  table[row sep=crcr]{%
1	0.16540853\\
2	0.30416488\\
3	0.36892283\\
4	0.43001318\\
};
%0.16540853 0.30416488 0.36892283 0.43001318
\addlegendentry{T=10}

\addplot [color=mycolor5, mark=asterisk, mark options={solid, mycolor5}, thick]
  table[row sep=crcr]{%
1	0.17686895\\
2	0.32740351\\
3	0.39470266\\
4	0.45144566\\
};
%0.17686895 0.32740351 0.39470266 0.45144566
\addlegendentry{T=20}

\end{axis}
\end{tikzpicture}% 
            \caption[Network2]%
            {{\small BERT / ERA}}
        \end{subfigure}
        \caption[ The average and standard deviation of critical parameters ]
        {Parameter analysis of InteGrad on SST2.} 
        \label{fig:integrad_sst2}
        \vspace{-4pt}
\end{figure}

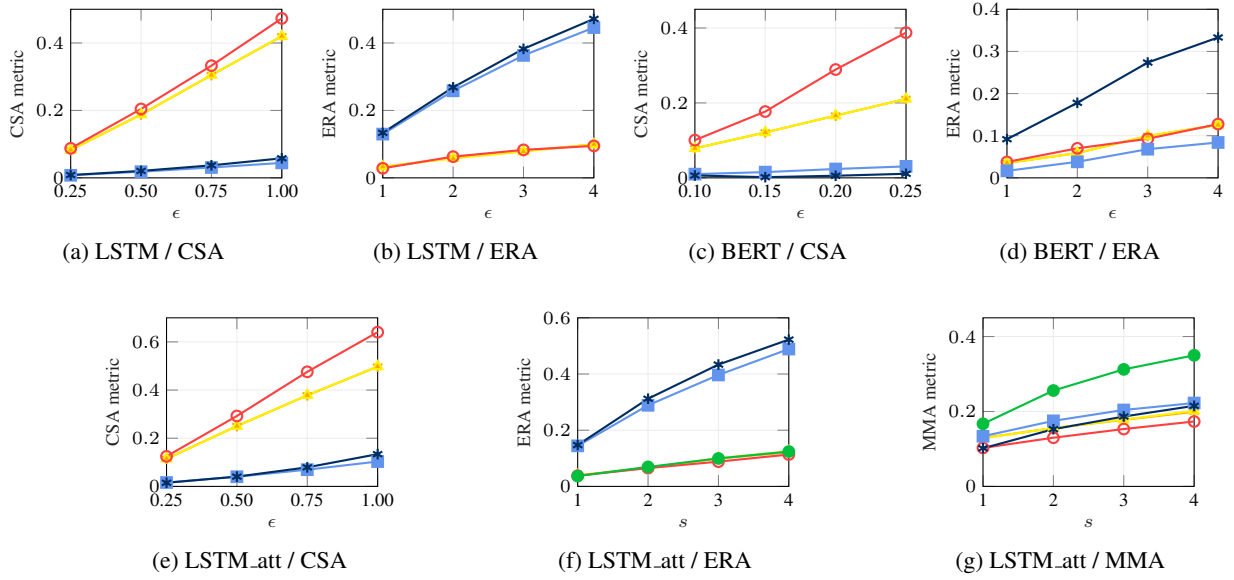
\begin{figure*}[t]
        \centering
        \begin{subfigure}[b]{0.23\textwidth}
        \setlength\figureheight{\heightunify in}
        \setlength\figurewidth{1.15in}
        \centering  \scriptsize
        % This file was created by matlab2tikz.
%
%The latest updates can be retrieved from
%  http://www.mathworks.com/matlabcentral/fileexchange/22022-matlab2tikz-matlab2tikz
%where you can also make suggestions and rate matlab2tikz.
%
\definecolor{mycolor1}{rgb}{1,0.65,0}%
\definecolor{mycolor2}{rgb}{1,0.94,0}%
\definecolor{mycolor3}{rgb}{1.0, 0.25, 0.25}%
\definecolor{mycolor4}{rgb}{0.39, 0.58, 0.93}
\definecolor{mycolor5}{rgb}{0., 0.18, 0.39}

\begin{tikzpicture}

\begin{axis}[%
width=0.951\figurewidth,
height=\figureheight,
at={(0\figurewidth,0\figureheight)},
scale only axis,
scaled x ticks=true,
xticklabels={0.25, 0.50, 0.75, 1.00},
xtick={1,2,3,4},
xmin=1,
xmax=4,
xlabel style={font=\color{white!15!black}},
xlabel={$\epsilon$},
ymin=0.00,
ymax=0.50,
grid, % --added
grid style={line width=.15pt, draw=gray!15}, % --added, dashed, 
ylabel style={font=\color{white!15!black}},
ylabel={CSA metric},
axis background/.style={fill=white},
axis background/.style={fill=white} % --added
]
\addplot [color=mycolor1, mark=asterisk, mark options={solid, mycolor1}, thick]
  table[row sep=crcr]{%
1	0.08266433752630502\\
2	0.18773215725146944\\
3	0.3047696851194565\\
4	0.42006351814296194\\
};
%0.08266433752630502,0.18773215725146944,0.3047696851194565,0.42006351814296194
%\addlegendentry{RAND}

\addplot [color=mycolor2, mark=triangle, mark options={solid, mycolor2}, thick]
  table[row sep=crcr]{%
1	0.08272602439637669\\
2	0.18787475967737877\\
3	0.30497163119880405\\
4	0.4202692513511436\\
};
%0.08272602439637669,0.18787475967737877,0.30497163119880405,0.4202692513511436
%\addlegendentry{SEP}

\addplot [color=mycolor3, mark=o, mark options={solid, mycolor3}, thick]
  table[row sep=crcr]{%
1	0.08758252876503336\\
2	0.20427776614528173\\
3	0.3327338874230015\\
4	0.4728740713306742\\
};
%0.08758252876503336,0.20427776614528173,0.3327338874230015,0.4728740713306742
%\addlegendentry{DFE}

\addplot [color=mycolor4, mark=square*, mark options={solid, mycolor4}, thick]
  table[row sep=crcr]{%
1	0.007598808947771159\\
2	0.01828011548903182\\
3	0.03078913539307395\\
4	0.04437813208497965\\
};
%0.007598808947771159,0.01828011548903182,0.03078913539307395,0.04437813208497965
%\addlegendentry{DFE}

\addplot [color=mycolor5, mark=asterisk, mark options={solid, mycolor5}, thick]
  table[row sep=crcr]{%
1	0.008422379759466324\\
2	0.020572135101285933\\
3	0.037094885240896235\\
4	0.058005073524164195\\
};
%0.008422379759466324,0.020572135101285933,0.037094885240896235,0.058005073524164195

\end{axis}
\end{tikzpicture}% 
            \caption[Network2]%
            {{\small LSTM / CSA}}
        \end{subfigure}
        \hfill
        \begin{subfigure}[b]{0.23\textwidth}
        \setlength\figureheight{\heightunify in}
        \setlength\figurewidth{1.15in}
        \centering  \scriptsize
        % This file was created by matlab2tikz.
%
%The latest updates can be retrieved from
%  http://www.mathworks.com/matlabcentral/fileexchange/22022-matlab2tikz-matlab2tikz
%where you can also make suggestions and rate matlab2tikz.
%
\definecolor{mycolor1}{rgb}{1,0.65,0}%
\definecolor{mycolor2}{rgb}{1,0.94,0}%
\definecolor{mycolor3}{rgb}{1.0, 0.25, 0.25}%
\definecolor{mycolor4}{rgb}{0.39, 0.58, 0.93}
\definecolor{mycolor5}{rgb}{0., 0.18, 0.39}

\begin{tikzpicture}

\begin{axis}[%
width=0.951\figurewidth,
height=\figureheight,
at={(0\figurewidth,0\figureheight)},
scale only axis,
scaled x ticks=true,
xticklabels={1,2,3,4},
xtick={1,2,3,4},
xmin=1,
xmax=4,
xlabel style={font=\color{white!15!black}},
xlabel={$\epsilon$},
ymin=0.00,
ymax=0.50,
grid, % --added
grid style={line width=.15pt, draw=gray!15}, % --added, dashed, 
ylabel style={font=\color{white!15!black}},
ylabel={ERA metric},
axis background/.style={fill=white},
axis background/.style={fill=white} % --added
]
\addplot [color=mycolor1, mark=asterisk, mark options={solid, mycolor1}, thick]
  table[row sep=crcr]{%
1	0.03203169\\
2	0.05969481\\
3	0.0796391\\
4	0.09926337\\
};
%0.03203169 0.05969481 0.0796391  0.09926337
%\addlegendentry{RAND}

\addplot [color=mycolor2, mark=triangle, mark options={solid, mycolor2}, thick]
  table[row sep=crcr]{%
1	0.03314192\\
2	0.05768058\\
3	0.07825523\\
4	0.09936075\\
};
%0.03314192 0.05768058 0.07825523 0.09936075
%\addlegendentry{SEP}

\addplot [color=mycolor3, mark=o, mark options={solid, mycolor3}, thick]
  table[row sep=crcr]{%
1	0.02833606\\
2	0.06296503\\
3	0.08306357\\
4	0.09475683\\
};
%0.02833606 0.06296503 0.08306357 0.09475683
%\addlegendentry{DFE}

\addplot [color=mycolor4, mark=square*, mark options={solid, mycolor4}, thick]
  table[row sep=crcr]{%
1	0.12976001\\
2	0.25809675\\
3	0.36313524\\
4	0.44577235\\
};
%0.12976001 0.25809675 0.36313524 0.44577235
%\addlegendentry{DFE}

\addplot [color=mycolor5, mark=asterisk, mark options={solid, mycolor5}, thick]
  table[row sep=crcr]{%
1	0.13358253\\
2	0.26846073\\
3	0.38264416\\
4	0.47150571\\
};
%0.13358253 0.26846073 0.38264416 0.47150571

\end{axis}
\end{tikzpicture}% 
            \caption[Network2]%
            {{\small LSTM / ERA}}
        \end{subfigure}
        \hfill
        %\vskip\baselineskip
        \begin{subfigure}[b]{0.23\textwidth}  
            \setlength\figureheight{\heightunify in}
        \setlength\figurewidth{1.15in}
        \centering  \scriptsize
        % This file was created by matlab2tikz.
%
%The latest updates can be retrieved from
%  http://www.mathworks.com/matlabcentral/fileexchange/22022-matlab2tikz-matlab2tikz
%where you can also make suggestions and rate matlab2tikz.
%
\definecolor{mycolor1}{rgb}{1,0.65,0}%
\definecolor{mycolor2}{rgb}{1,0.94,0}%
\definecolor{mycolor3}{rgb}{1.0, 0.25, 0.25}%
\definecolor{mycolor4}{rgb}{0.39, 0.58, 0.93}
\definecolor{mycolor5}{rgb}{0., 0.18, 0.39}

\begin{tikzpicture}

\begin{axis}[%
width=0.951\figurewidth,
height=\figureheight,
at={(0\figurewidth,0\figureheight)},
scale only axis,
scaled x ticks=true,
xticklabels={0.10, 0.15, 0.20, 0.25},
xtick={1,2,3,4},
xmin=1,
xmax=4,
xlabel style={font=\color{white!15!black}},
xlabel={$\epsilon$},
ymin=0.00,
ymax=0.45,
grid, % --added
grid style={line width=.15pt, draw=gray!15}, % --added, dashed, 
ylabel style={font=\color{white!15!black}},
ylabel={CSA metric},
axis background/.style={fill=white},
axis background/.style={fill=white} % --added
]
\addplot [color=mycolor1, mark=asterisk, mark options={solid, mycolor1}, thick]
  table[row sep=crcr]{%
1	0.07879502868818769\\
2	0.12120331027922458\\
3	0.1661627616636369\\
4	0.21072369824214163\\
};
%0.07879502868818769,0.12120331027922458,0.1661627616636369,0.21072369824214163
%\addlegendentry{RAND}

\addplot [color=mycolor2, mark=triangle, mark options={solid, mycolor2}, thick]
  table[row sep=crcr]{%
1	0.07879380416222746\\
2	0.12119366012921327\\
3	0.16615605201896713\\
4	0.21071859219641545\\
};
%0.07879380416222746,0.12119366012921327,0.16615605201896713,0.21071859219641545
%\addlegendentry{SEP}

\addplot [color=mycolor3, mark=o, mark options={solid, mycolor3}, thick]
  table[row sep=crcr]{%
1	0.10071931750319664\\
2	0.17699492833477373\\
3	0.28913116189292126\\
4	0.3875843816903488\\
};
%0.10071931750319664,0.17699492833477373,0.28913116189292126,0.3875843816903488
%\addlegendentry{DFE}

\addplot [color=mycolor4, mark=square*, mark options={solid, mycolor4}, thick]
  table[row sep=crcr]{%
1	0.009927852337568993\\
2	0.015377943015428067\\
3	0.023618519562983387\\
4	0.030649219437196625\\
};
%0.009927852337568993,0.015377943015428067,0.023618519562983387,0.030649219437196625
%\addlegendentry{DFE}

\addplot [color=mycolor5, mark=asterisk, mark options={solid, mycolor5}, thick]
  table[row sep=crcr]{%
1	0.006405343597418248\\
2	0.0018324167904487862\\
3	0.005633576654425725\\
4	0.010902105977944243\\
};
%0.006405343597418248,0.0018324167904487862,0.005633576654425725,0.010902105977944243

\end{axis}
\end{tikzpicture}% 
            \caption[]%
            {{\small BERT / CSA}}
        \end{subfigure}
        \hfill
        \begin{subfigure}[b]{0.23\textwidth}  
            \setlength\figureheight{\heightunify in}
        \setlength\figurewidth{1.15in}
        \centering  \scriptsize
        % This file was created by matlab2tikz.
%
%The latest updates can be retrieved from
%  http://www.mathworks.com/matlabcentral/fileexchange/22022-matlab2tikz-matlab2tikz
%where you can also make suggestions and rate matlab2tikz.
%
\definecolor{mycolor1}{rgb}{1,0.65,0}%
\definecolor{mycolor2}{rgb}{1,0.94,0}%
\definecolor{mycolor3}{rgb}{1.0, 0.25, 0.25}%
\definecolor{mycolor4}{rgb}{0.39, 0.58, 0.93}
\definecolor{mycolor5}{rgb}{0., 0.18, 0.39}

\begin{tikzpicture}

\begin{axis}[%
width=0.951\figurewidth,
height=\figureheight,
at={(0\figurewidth,0\figureheight)},
scale only axis,
scaled x ticks=true,
xticklabels={1,2,3,4},
xtick={1,2,3,4},
xmin=1,
xmax=4,
xlabel style={font=\color{white!15!black}},
xlabel={$\epsilon$},
ymin=0.00,
ymax=0.40,
grid, % --added
grid style={line width=.15pt, draw=gray!15}, % --added, dashed, 
ylabel style={font=\color{white!15!black}},
ylabel={ERA metric},
axis background/.style={fill=white},
axis background/.style={fill=white} % --added
]
\addplot [color=mycolor1, mark=asterisk, mark options={solid, mycolor1}, thick]
  table[row sep=crcr]{%
1	0.03521278\\
2	0.05883501\\
3	0.09926682\\
4	0.12553168\\
};
%0.03521278 0.05883501 0.09926682 0.12553168
%\addlegendentry{RAND}

\addplot [color=mycolor2, mark=triangle, mark options={solid, mycolor2}, thick]
  table[row sep=crcr]{%
1	0.03521297\\
2	0.05883501\\
3	0.09941073\\
4	0.12553275\\
};
%0.03521297 0.05883501 0.09941073 0.12553275
%\addlegendentry{SEP}

\addplot [color=mycolor3, mark=o, mark options={solid, mycolor3}, thick]
  table[row sep=crcr]{%
1	0.03774742\\
2	0.07008029\\
3	0.09261512\\
4	0.12762946\\
};
%0.03774742 0.07008029 0.09261512 0.12762946
%\addlegendentry{DFE}

\addplot [color=mycolor4, mark=square*, mark options={solid, mycolor4}, thick]
  table[row sep=crcr]{%
1	0.016594\\
2	0.03806318\\
3	0.06805319\\
4	0.08392287\\
};
%0.016594   0.03806318 0.06805319 0.08392287
%\addlegendentry{DFE}

\addplot [color=mycolor5, mark=asterisk, mark options={solid, mycolor5}, thick]
  table[row sep=crcr]{%
1	0.09168428\\
2	0.17803603\\
3	0.27358104\\
4	0.33309593\\
};
%0.09168428 0.17803603 0.27358104 0.33309593

\end{axis}
\end{tikzpicture}% 
            \caption[]%
            {{\small BERT / ERA}}
        \end{subfigure}

\vskip\baselineskip

        \hfill
        \begin{subfigure}[b]{0.25\textwidth}
        \setlength\figureheight{\heightunify in}
        \setlength\figurewidth{1.15in}
        \centering  \scriptsize
        % This file was created by matlab2tikz.
%
%The latest updates can be retrieved from
%  http://www.mathworks.com/matlabcentral/fileexchange/22022-matlab2tikz-matlab2tikz
%where you can also make suggestions and rate matlab2tikz.
%
\definecolor{mycolor1}{rgb}{1,0.65,0}%
\definecolor{mycolor2}{rgb}{1,0.94,0}%
\definecolor{mycolor3}{rgb}{1.0, 0.25, 0.25}%
\definecolor{mycolor4}{rgb}{0.39, 0.58, 0.93}
\definecolor{mycolor5}{rgb}{0., 0.18, 0.39}

\begin{tikzpicture}

\begin{axis}[%
width=0.951\figurewidth,
height=\figureheight,
at={(0\figurewidth,0\figureheight)},
scale only axis,
scaled x ticks=true,
xticklabels={0.25, 0.50, 0.75, 1.00},
xtick={1,2,3,4},
xmin=1,
xmax=4,
xlabel style={font=\color{white!15!black}},
xlabel={$\epsilon$},
ymin=0.00,
ymax=0.70,
grid, % --added
grid style={line width=.15pt, draw=gray!15}, % --added, dashed, 
ylabel style={font=\color{white!15!black}},
ylabel={CSA metric},
axis background/.style={fill=white},
axis background/.style={fill=white} % --added
]
\addplot [color=mycolor1, mark=asterisk, mark options={solid, mycolor1}, thick]
  table[row sep=crcr]{%
1	0.11468997061186843\\
2	0.250487032538059\\
3	0.3782089415552371\\
4	0.49722286757846956\\
};
%0.11468997061186843,0.250487032538059,0.3782089415552371,0.49722286757846956
%\addlegendentry{RAND}

\addplot [color=mycolor2, mark=triangle, mark options={solid, mycolor2}, thick]
  table[row sep=crcr]{%
1	0.11476968268377967\\
2	0.25079558379292277\\
3	0.37878659839814827\\
4	0.49795466986958864\\
};
%0.11476968268377967,0.25079558379292277,0.37878659839814827,0.49795466986958864
%\addlegendentry{SEP}

\addplot [color=mycolor3, mark=o, mark options={solid, mycolor3}, thick]
  table[row sep=crcr]{%
1	0.12478121114174977\\
2	0.29238197928486187\\
3	0.4757432031002352\\
4	0.640452080854777\\
};
%0.12478121114174977,0.29238197928486187,0.4757432031002352,0.640452080854777
%\addlegendentry{DFE}

\addplot [color=mycolor4, mark=square*, mark options={solid, mycolor4}, thick]
  table[row sep=crcr]{%
1	0.016177733352720648\\
2	0.040135994156754895\\
3	0.06945451361336799\\
4	0.10318444802923755\\
};
%0.016177733352720648,0.040135994156754895,0.06945451361336799,0.10318444802923755
%\addlegendentry{DFE}

\addplot [color=mycolor5, mark=asterisk, mark options={solid, mycolor5}, thick]
  table[row sep=crcr]{%
1	0.015031219346353769\\
2	0.0406084366705304\\
3	0.07913584126637442\\
4	0.1338555717488969\\
};
%0.015031219346353769,0.0406084366705304,0.07913584126637442,0.1338555717488969

\end{axis}
\end{tikzpicture}% 
            \caption[Network2]%
            {{\small LSTM\_att / CSA}}
        \end{subfigure}
        \hfill
        \begin{subfigure}[b]{0.25\textwidth}  
            \setlength\figureheight{\heightunify in}
        \setlength\figurewidth{1.15in}
        \centering  \scriptsize
        % This file was created by matlab2tikz.
%
%The latest updates can be retrieved from
%  http://www.mathworks.com/matlabcentral/fileexchange/22022-matlab2tikz-matlab2tikz
%where you can also make suggestions and rate matlab2tikz.
%
\definecolor{mycolor1}{rgb}{1,0.65,0}%
\definecolor{mycolor2}{rgb}{1,0.94,0}%
\definecolor{mycolor3}{rgb}{1.0, 0.25, 0.25}%
\definecolor{mycolor4}{rgb}{0.39, 0.58, 0.93}
\definecolor{mycolor5}{rgb}{0., 0.18, 0.39}
\definecolor{mycolor6}{rgb}{0.01, 0.75, 0.24}

\begin{tikzpicture}

\begin{axis}[%
width=0.951\figurewidth,
height=\figureheight,
at={(0\figurewidth,0\figureheight)},
scale only axis,
scaled x ticks=true,
xticklabels={1,2,3,4},
xtick={1,2,3,4},
xmin=1,
xmax=4,
xlabel style={font=\color{white!15!black}},
xlabel={$s$},
ymin=0.00,
ymax=0.60,
grid, % --added
grid style={line width=.15pt, draw=gray!15}, % --added, dashed, 
ylabel style={font=\color{white!15!black}},
ylabel={ERA metric},
axis background/.style={fill=white},
axis background/.style={fill=white} % --added
]
\addplot [color=mycolor1, mark=asterisk, mark options={solid, mycolor1}, thick]
  table[row sep=crcr]{%
1	0.03923316\\
2	0.06666047\\
3	0.09965845\\
4	0.12095924\\
};
%0.03923316 0.06666047 0.09965845 0.12095924
%\addlegendentry{RAND}

\addplot [color=mycolor2, mark=triangle, mark options={solid, mycolor2}, thick]
  table[row sep=crcr]{%
1	0.03922813\\
2	0.0660146\\
3	0.10015032\\
4	0.12481962\\
};
%0.03922813 0.0660146  0.10015032 0.12481962
%\addlegendentry{SEP}

\addplot [color=mycolor3, mark=o, mark options={solid, mycolor3}, thick]
  table[row sep=crcr]{%
1	0.03798298\\
2	0.06511595\\
3	0.08815127\\
4	0.11345229\\
};
%0.03798298 0.06511595 0.08815127 0.11345229
%\addlegendentry{DFE}

\addplot [color=mycolor4, mark=square*, mark options={solid, mycolor4}, thick]
  table[row sep=crcr]{%
1	0.14430331\\
2	0.28875522\\
3	0.39662543\\
4	0.48892653\\
};
%0.14430331 0.28875522 0.39662543 0.48892653
%\addlegendentry{DFE}

\addplot [color=mycolor5, mark=asterisk, mark options={solid, mycolor5}, thick]
  table[row sep=crcr]{%
1	0.14716823\\
2	0.31212834\\
3	0.4336269\\
4	0.52257752\\
};
%0.14716823 0.31212834 0.4336269  0.52257752

\addplot [color=mycolor6, mark=*, mark options={solid, mycolor6}, thick]
  table[row sep=crcr]{%
1	0.03678363\\
2	0.06930974\\
3	0.0995075\\
4	0.12405418\\
};
%0.03678363 0.06930974 0.0995075  0.12405418

\end{axis}
\end{tikzpicture}% 
            \caption[]%
            {{\small LSTM\_att / ERA}}
        \end{subfigure}
        \hfill
        \begin{subfigure}[b]{0.25\textwidth}  
            \setlength\figureheight{\heightunify in}
        \setlength\figurewidth{1.15in}
        \centering  \scriptsize
        % This file was created by matlab2tikz.
%
%The latest updates can be retrieved from
%  http://www.mathworks.com/matlabcentral/fileexchange/22022-matlab2tikz-matlab2tikz
%where you can also make suggestions and rate matlab2tikz.
%
\definecolor{mycolor1}{rgb}{1,0.65,0}%
\definecolor{mycolor2}{rgb}{1,0.94,0}%
\definecolor{mycolor3}{rgb}{1.0, 0.25, 0.25}%
\definecolor{mycolor4}{rgb}{0.39, 0.58, 0.93}
\definecolor{mycolor5}{rgb}{0., 0.18, 0.39}
\definecolor{mycolor6}{rgb}{0.01, 0.75, 0.24}

\begin{tikzpicture}

\begin{axis}[%
width=0.951\figurewidth,
height=\figureheight,
at={(0\figurewidth,0\figureheight)},
scale only axis,
scaled x ticks=true,
xticklabels={1,2,3,4},
xtick={1,2,3,4},
xmin=1,
xmax=4,
xlabel style={font=\color{white!15!black}},
xlabel={$s$},
ymin=0.00,
ymax=0.45,
grid, % --added
grid style={line width=.15pt, draw=gray!15}, % --added, dashed, 
ylabel style={font=\color{white!15!black}},
ylabel={MMA metric},
axis background/.style={fill=white},
axis background/.style={fill=white} % --added
]
\addplot [color=mycolor1, mark=asterisk, mark options={solid, mycolor1}, thick]
  table[row sep=crcr]{%
1	0.12935184\\
2	0.15563498\\
3	0.1776591\\
4	0.19901004\\
};
%0.12935184 0.15563498 0.1776591  0.19901004
%\addlegendentry{RAND}

\addplot [color=mycolor2, mark=triangle, mark options={solid, mycolor2}, thick]
  table[row sep=crcr]{%
1	0.12855456\\
2	0.15592351\\
3	0.17836879\\
4	0.20177354\\
};
%0.12855456 0.15592351 0.17836879 0.20177354
%\addlegendentry{SEP}

\addplot [color=mycolor3, mark=o, mark options={solid, mycolor3}, thick]
  table[row sep=crcr]{%
1	0.10269851\\
2	0.12994292\\
3	0.1531387\\
4	0.17321617\\
};
%0.10269851 0.12994292 0.1531387  0.17321617
%\addlegendentry{DFE}

\addplot [color=mycolor4, mark=square*, mark options={solid, mycolor4}, thick]
  table[row sep=crcr]{%
1	0.13388467\\
2	0.17494437\\
3	0.20402395\\
4	0.22253873\\
};
%0.13388467 0.17494437 0.20402395 0.22253873
%\addlegendentry{DFE}

\addplot [color=mycolor5, mark=asterisk, mark options={solid, mycolor5}, thick]
  table[row sep=crcr]{%
1	0.10183281\\
2	0.15262763\\
3	0.18654279\\
4	0.21507918\\
};
%0.10183281 0.15262763 0.18654279 0.21507918

\addplot [color=mycolor6, mark=*, mark options={solid, mycolor6}, thick]
  table[row sep=crcr]{%
1	0.16702702\\
2	0.25587019\\
3	0.31236272\\
4	0.34992869\\
};
%0.16702702 0.25587019 0.31236272 0.34992869

\end{axis}
\end{tikzpicture}% 
            \caption[]%
            {{\small LSTM\_att / MMA}}
        \end{subfigure}
        \hfill
        \caption[ The average and standard deviation of critical parameters ]
        {Interpretation faithfulness comparison under different metrics on Yelp dataset. \vagrad:VaGrad,\, \smoothgrad:SmoothGrad,\, \itergrad:IterGrad,\, \inpgrad:InpGrad,\, \integrad:InteGrad,\, \rankmask:RankMask.} 
        \label{fig:yelp_cross}
        \vspace{-0pt}
\end{figure*}

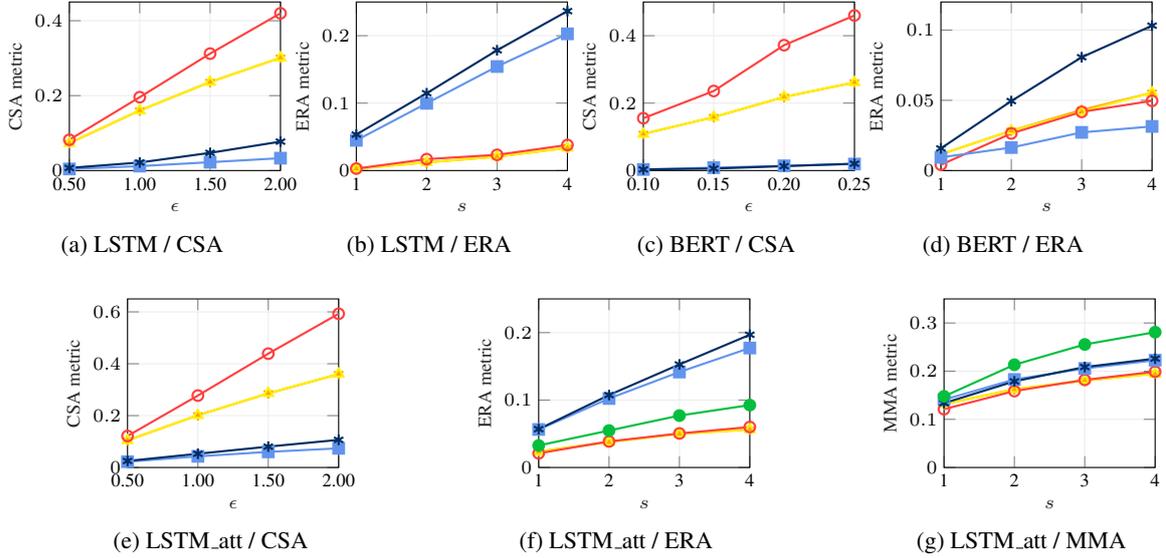
\begin{figure*}[h!]
        \centering
        \begin{subfigure}[b]{0.23\textwidth}
        \setlength\figureheight{\heightunify in}
        \setlength\figurewidth{1.15in}
        \centering  \scriptsize
        % This file was created by matlab2tikz.
%
%The latest updates can be retrieved from
%  http://www.mathworks.com/matlabcentral/fileexchange/22022-matlab2tikz-matlab2tikz
%where you can also make suggestions and rate matlab2tikz.
%
\definecolor{mycolor1}{rgb}{1,0.65,0}%
\definecolor{mycolor2}{rgb}{1,0.94,0}%
\definecolor{mycolor3}{rgb}{1.0, 0.25, 0.25}%
\definecolor{mycolor4}{rgb}{0.39, 0.58, 0.93}
\definecolor{mycolor5}{rgb}{0., 0.18, 0.39}

\begin{tikzpicture}

\begin{axis}[%
width=0.951\figurewidth,
height=\figureheight,
at={(0\figurewidth,0\figureheight)},
scale only axis,
scaled x ticks=true,
xticklabels={0.50, 1.00, 1.50, 2.00},
xtick={1,2,3,4},
xmin=1,
xmax=4,
xlabel style={font=\color{white!15!black}},
xlabel={$\epsilon$},
ymin=0.00,
ymax=0.45,
grid, % --added
grid style={line width=.15pt, draw=gray!15}, % --added, dashed, 
ylabel style={font=\color{white!15!black}},
ylabel={CSA metric},
axis background/.style={fill=white},
axis background/.style={fill=white} % --added
]
\addplot [color=mycolor1, mark=asterisk, mark options={solid, mycolor1}, thick]
  table[row sep=crcr]{%
1	0.07305827037978378\\
2	0.16031299244619276\\
3	0.2362683952427387\\
4	0.30115392162252214\\
};
% 0.07305827037978378,0.16031299244619276,0.2362683952427387,0.30115392162252214
%\addlegendentry{RAND}

\addplot [color=mycolor2, mark=triangle, mark options={solid, mycolor2}, thick]
  table[row sep=crcr]{%
1	0.0730877866770889\\
2	0.16039213050132242\\
3	0.23636695409946015\\
4	0.3012912954745199\\
};
% 0.0730877866770889,0.16039213050132242,0.23636695409946015,0.3012912954745199
%\addlegendentry{SEP}

\addplot [color=mycolor3, mark=o, mark options={solid, mycolor3}, thick]
  table[row sep=crcr]{%
1	0.08224613106020298\\
2	0.1958429034865814\\
3	0.3121543229212044\\
4	0.4197465257633811\\
};
% 0.0730877866770889,0.16039213050132242,0.23636695409946015,0.3012912954745199
%\addlegendentry{DFE}

\addplot [color=mycolor4, mark=square*, mark options={solid, mycolor4}, thick]
  table[row sep=crcr]{%
1	0.0042334924949356135\\
2	0.01183801507377255\\
3	0.022341151172245395\\
4	0.033063135660936376\\
};
% 0.0042334924949356135,0.01183801507377255,0.022341151172245395,0.033063135660936376
%\addlegendentry{DFE}

\addplot [color=mycolor5, mark=asterisk, mark options={solid, mycolor5}, thick]
  table[row sep=crcr]{%
1	0.007060071320789967\\
2	0.02150833457066819\\
3	0.04726741479447462\\
4	0.07727407398200065\\
};
% 0.007060071320789967,0.02150833457066819,0.04726741479447462,0.07727407398200065
\end{axis}
\end{tikzpicture}% 
            \caption[Network2]%
            {{\small LSTM / CSA}}
        \end{subfigure}
        %\hfill
        \begin{subfigure}[b]{0.23\textwidth}
        \setlength\figureheight{\heightunify in}
        \setlength\figurewidth{1.15in}
        \centering  \scriptsize
        % This file was created by matlab2tikz.
%
%The latest updates can be retrieved from
%  http://www.mathworks.com/matlabcentral/fileexchange/22022-matlab2tikz-matlab2tikz
%where you can also make suggestions and rate matlab2tikz.
%
\definecolor{mycolor1}{rgb}{1,0.65,0}%
\definecolor{mycolor2}{rgb}{1,0.94,0}%
\definecolor{mycolor3}{rgb}{1.0, 0.25, 0.25}%
\definecolor{mycolor4}{rgb}{0.39, 0.58, 0.93}
\definecolor{mycolor5}{rgb}{0., 0.18, 0.39}

\begin{tikzpicture}

\begin{axis}[%
width=0.951\figurewidth,
height=\figureheight,
at={(0\figurewidth,0\figureheight)},
scale only axis,
scaled x ticks=true,
xticklabels={1,2,3,4},
xtick={1,2,3,4},
xmin=1,
xmax=4,
xlabel style={font=\color{white!15!black}},
xlabel={$s$},
ymin=0.00,
ymax=0.25,
grid, % --added
grid style={line width=.15pt, draw=gray!15}, % --added, dashed, 
ylabel style={font=\color{white!15!black}},
ylabel={ERA metric},
axis background/.style={fill=white},
axis background/.style={fill=white} % --added
]
\addplot [color=mycolor1, mark=asterisk, mark options={solid, mycolor1}, thick]
  table[row sep=crcr]{%
1	0.00282758\\
2	0.012077\\
3	0.02055144\\
4	0.03366198\\
};
%0.00282758 0.012077   0.02055144 0.03366198
%\addlegendentry{VaGrad}

\addplot [color=mycolor2, mark=triangle, mark options={solid, mycolor2}, thick]
  table[row sep=crcr]{%
1	0.00278548\\
2	0.01209439\\
3	0.02071838\\
4	0.03355474\\
};
%0.00278548 0.01209439 0.02071838 0.03355474
%\addlegendentry{SmoothGrad}

\addplot [color=mycolor3, mark=o, mark options={solid, mycolor3}, thick]
  table[row sep=crcr]{%
1	0.00294248\\
2	0.01695026\\
3	0.0233514\\
4	0.03795785\\
};
%0.00294248 0.01695026 0.0233514  0.03795785
%\addlegendentry{IterGrad}

\addplot [color=mycolor4, mark=square*, mark options={solid, mycolor4}, thick]
  table[row sep=crcr]{%
1	0.04481789\\
2	0.09971746\\
3	0.15429465\\
4	0.20280115\\
};
%0.04481789 0.09971746 0.15429465 0.20280115
%\addlegendentry{InpGrad}

\addplot [color=mycolor5, mark=asterisk, mark options={solid, mycolor5}, thick]
  table[row sep=crcr]{%
1	0.05339326\\
2	0.11468772\\
3	0.178396\\
4	0.23643169\\
};
%0.05339326 0.11468772 0.178396   0.23643169
%\addlegendentry{InteGrad}

\end{axis}
\end{tikzpicture}%

% [0.00282758 0.012077   0.02055144 0.03366198]
% [0.00294248 0.01695026 0.0233514  0.03795785]
% [0.00278548 0.01209439 0.02071838 0.03355474]
% [0.04481789 0.09971746 0.15429465 0.20280115]
% [0.05339326 0.11468772 0.178396   0.23643169] 
            \caption[Network2]%
            {{\small LSTM / ERA}}
        \end{subfigure}
        %\hfill
        %\vskip\baselineskip
        \begin{subfigure}[b]{0.23\textwidth}  
            \setlength\figureheight{\heightunify in}
        \setlength\figurewidth{1.15in}
        \centering  \scriptsize
        % This file was created by matlab2tikz.
%
%The latest updates can be retrieved from
%  http://www.mathworks.com/matlabcentral/fileexchange/22022-matlab2tikz-matlab2tikz
%where you can also make suggestions and rate matlab2tikz.
%
\definecolor{mycolor1}{rgb}{1,0.65,0}%
\definecolor{mycolor2}{rgb}{1,0.94,0}%
\definecolor{mycolor3}{rgb}{1.0, 0.25, 0.25}%
\definecolor{mycolor4}{rgb}{0.39, 0.58, 0.93}
\definecolor{mycolor5}{rgb}{0., 0.18, 0.39}

\begin{tikzpicture}

\begin{axis}[%
width=0.951\figurewidth,
height=\figureheight,
at={(0\figurewidth,0\figureheight)},
scale only axis,
scaled x ticks=true,
xticklabels={0.10, 0.15, 0.20, 0.25},
xtick={1,2,3,4},
xmin=1,
xmax=4,
xlabel style={font=\color{white!15!black}},
xlabel={$\epsilon$},
ymin=0.00,
ymax=0.50,
grid, % --added
grid style={line width=.15pt, draw=gray!15}, % --added, dashed, 
ylabel style={font=\color{white!15!black}},
ylabel={CSA metric},
axis background/.style={fill=white},
axis background/.style={fill=white} % --added
]
\addplot [color=mycolor1, mark=asterisk, mark options={solid, mycolor1}, thick]
  table[row sep=crcr]{%
1	0.10853785803078511\\
2	0.15850103406671337\\
3	0.21828132244297388\\
4	0.2613348546478001\\
};
%0.10853785803078511,0.15850103406671337,0.21828132244297388,0.2613348546478001
%\addlegendentry{RAND}

\addplot [color=mycolor2, mark=triangle, mark options={solid, mycolor2}, thick]
  table[row sep=crcr]{%
1	0.10855510733028317\\
2	0.15851269657953382\\
3	0.21852918934569834\\
4	0.26180337905033335\\
};
%0.10855510733028317,0.15851269657953382,0.21852918934569834,0.26180337905033335
%\addlegendentry{SEP}

\addplot [color=mycolor3, mark=o, mark options={solid, mycolor3}, thick]
  table[row sep=crcr]{%
1	0.15523499263565232\\
2	0.23589980207353548\\
3	0.3718750560548432\\
4	0.4599740230492674\\
};
%0.15523499263565232,0.23589980207353548,0.3718750560548432,0.4599740230492674
%\addlegendentry{DFE}

\addplot [color=mycolor4, mark=square*, mark options={solid, mycolor4}, thick]
  table[row sep=crcr]{%
1	0.003201617410570195\\
2	0.008673299300962278\\
3	0.014099401868524236\\
4	0.019350811807523384\\
};
%0.003201617410570195,0.008673299300962278,0.014099401868524236,0.019350811807523384
%\addlegendentry{DFE}

\addplot [color=mycolor5, mark=asterisk, mark options={solid, mycolor5}, thick]
  table[row sep=crcr]{%
1	0.0034033620949210277\\
2	0.0059370211735448415\\
3	0.01308889271831953\\
4	0.02041085179183491\\
};
%0.0034033620949210277,0.0059370211735448415,0.01308889271831953,0.02041085179183491

\end{axis}
\end{tikzpicture}% 
            \caption[]%
            {{\small BERT / CSA}}
        \end{subfigure}
        %\hfill
        \begin{subfigure}[b]{0.23\textwidth}  
            \setlength\figureheight{\heightunify in}
        \setlength\figurewidth{1.15in}
        \centering  \scriptsize
        % This file was created by matlab2tikz.
%
%The latest updates can be retrieved from
%  http://www.mathworks.com/matlabcentral/fileexchange/22022-matlab2tikz-matlab2tikz
%where you can also make suggestions and rate matlab2tikz.
%
\definecolor{mycolor1}{rgb}{1,0.65,0}%
\definecolor{mycolor2}{rgb}{1,0.94,0}%
\definecolor{mycolor3}{rgb}{1.0, 0.25, 0.25}%
\definecolor{mycolor4}{rgb}{0.39, 0.58, 0.93}
\definecolor{mycolor5}{rgb}{0., 0.18, 0.39}

\begin{tikzpicture}

\begin{axis}[%
width=0.951\figurewidth,
height=\figureheight,
at={(0\figurewidth,0\figureheight)},
scale only axis,
scaled x ticks=true,
xticklabels={1,2,3,4},
xtick={1,2,3,4},
xmin=1,
xmax=4,
xlabel style={font=\color{white!15!black}},
xlabel={$s$},
ymin=0.00,
ymax=0.12,
yticklabel style={
        /pgf/number format/fixed,
        /pgf/number format/precision=3
},
grid, % --added
grid style={line width=.15pt, draw=gray!15}, % --added, dashed, 
ylabel style={font=\color{white!15!black}},
ylabel={ERA metric},
axis background/.style={fill=white},
axis background/.style={fill=white} % --added
]
\addplot [color=mycolor1, mark=asterisk, mark options={solid, mycolor1}, thick]
  table[row sep=crcr]{%
1	0.0115331\\
2	0.02836552\\
3	0.04315841\\
4	0.0553902\\
};
%0.0115331  0.02836552 0.04315841 0.0553902
%\addlegendentry{RAND}

\addplot [color=mycolor2, mark=triangle, mark options={solid, mycolor2}, thick]
  table[row sep=crcr]{%
1	0.0115351\\
2	0.02836552\\
3	0.041908\\
4	0.05538826\\
};
%0.0115351  0.02836552 0.041908   0.05538826
%\addlegendentry{SEP}

\addplot [color=mycolor3, mark=o, mark options={solid, mycolor3}, thick]
  table[row sep=crcr]{%
1	0.00412363\\
2	0.02641783\\
3	0.04175649\\
4	0.04963938\\
};
%0.00412363 0.02641783 0.04175649 0.04963938
%\addlegendentry{DFE}

\addplot [color=mycolor4, mark=square*, mark options={solid, mycolor4}, thick]
  table[row sep=crcr]{%
1	0.00939508\\
2	0.01637094\\
3	0.02716969\\
4	0.03140361\\
};
%0.00939508 0.01637094 0.02716969 0.03140361
%\addlegendentry{DFE}

\addplot [color=mycolor5, mark=asterisk, mark options={solid, mycolor5}, thick]
  table[row sep=crcr]{%
1	0.01579929\\
2	0.04946456\\
3	0.08065706\\
4	0.10314986\\
};
%0.01579929 0.04946456 0.08065706 0.10314986

\end{axis}
\end{tikzpicture}% 
            \caption[]%
            {{\small BERT / ERA}}
        \end{subfigure}

\vskip\baselineskip

        \hfill
        \begin{subfigure}[b]{0.25\textwidth}
        \setlength\figureheight{\heightunify in}
        \setlength\figurewidth{1.15in}
        \centering  \scriptsize
        % This file was created by matlab2tikz.
%
%The latest updates can be retrieved from
%  http://www.mathworks.com/matlabcentral/fileexchange/22022-matlab2tikz-matlab2tikz
%where you can also make suggestions and rate matlab2tikz.
%
\definecolor{mycolor1}{rgb}{1,0.65,0}%
\definecolor{mycolor2}{rgb}{1,0.94,0}%
\definecolor{mycolor3}{rgb}{1.0, 0.25, 0.25}%
\definecolor{mycolor4}{rgb}{0.39, 0.58, 0.93}
\definecolor{mycolor5}{rgb}{0., 0.18, 0.39}

\begin{tikzpicture}

\begin{axis}[%
width=0.951\figurewidth,
height=\figureheight,
at={(0\figurewidth,0\figureheight)},
scale only axis,
scaled x ticks=true,
xticklabels={0.50, 1.00, 1.50, 2.00},
xtick={1,2,3,4},
xmin=1,
xmax=4,
xlabel style={font=\color{white!15!black}},
xlabel={$\epsilon$},
ymin=0.00,
ymax=0.65,
grid, % --added
grid style={line width=.15pt, draw=gray!15}, % --added, dashed, 
ylabel style={font=\color{white!15!black}},
ylabel={CSA metric},
axis background/.style={fill=white},
axis background/.style={fill=white} % --added
]
\addplot [color=mycolor1, mark=asterisk, mark options={solid, mycolor1}, thick]
  table[row sep=crcr]{%
1	0.10454040796534377\\
2	0.20121729340435082\\
3	0.2858113334465498\\
4	0.360112844601835\\
};
%0.10454040796534377,0.20121729340435082,0.2858113334465498,0.360112844601835
%\addlegendentry{RAND}

\addplot [color=mycolor2, mark=triangle, mark options={solid, mycolor2}, thick]
  table[row sep=crcr]{%
1	0.10471260307400043\\
2	0.2015873667109831\\
3	0.28643321165188795\\
4	0.36088317272116566\\
};
%0.10471260307400043,0.2015873667109831,0.28643321165188795,0.36088317272116566
%\addlegendentry{SEP}

\addplot [color=mycolor3, mark=o, mark options={solid, mycolor3}, thick]
  table[row sep=crcr]{%
1	0.1214627213382613\\
2	0.2773023154104732\\
3	0.4389133243603435\\
4	0.5927685390975879\\
};
%0.1214627213382613,0.2773023154104732,0.4389133243603435,0.5927685390975879
%\addlegendentry{DFE}

\addplot [color=mycolor4, mark=square*, mark options={solid, mycolor4}, thick]
  table[row sep=crcr]{%
1	0.02222195419967961\\
2	0.04294598297462513\\
3	0.06022225904288401\\
4	0.07384509850206511\\
};
%0.02222195419967961,0.04294598297462513,0.06022225904288401,0.07384509850206511
%\addlegendentry{DFE}

\addplot [color=mycolor5, mark=asterisk, mark options={solid, mycolor5}, thick]
  table[row sep=crcr]{%
1	0.02517825838252115\\
2	0.052857632096749034\\
3	0.08044833818696831\\
4	0.1065357363809332\\
};
%0.02517825838252115,0.052857632096749034,0.08044833818696831,0.1065357363809332
\end{axis}
\end{tikzpicture}% 
            \caption[Network2]%
            {{\small LSTM\_att / CSA}}
        \end{subfigure}
        \hfill
        \begin{subfigure}[b]{0.25\textwidth}  
            \setlength\figureheight{\heightunify in}
        \setlength\figurewidth{1.15in}
        \centering  \scriptsize
        % This file was created by matlab2tikz.
%
%The latest updates can be retrieved from
%  http://www.mathworks.com/matlabcentral/fileexchange/22022-matlab2tikz-matlab2tikz
%where you can also make suggestions and rate matlab2tikz.
%
\definecolor{mycolor1}{rgb}{1,0.65,0}%
\definecolor{mycolor2}{rgb}{1,0.94,0}%
\definecolor{mycolor3}{rgb}{1.0, 0.25, 0.25}%
\definecolor{mycolor4}{rgb}{0.39, 0.58, 0.93}
\definecolor{mycolor5}{rgb}{0., 0.18, 0.39}
\definecolor{mycolor6}{rgb}{0.01, 0.75, 0.24}

\begin{tikzpicture}

\begin{axis}[%
width=0.951\figurewidth,
height=\figureheight,
at={(0\figurewidth,0\figureheight)},
scale only axis,
scaled x ticks=true,
xticklabels={1,2,3,4},
xtick={1,2,3,4},
xmin=1,
xmax=4,
xlabel style={font=\color{white!15!black}},
xlabel={$s$},
ymin=0.00,
ymax=0.25,
grid, % --added
grid style={line width=.15pt, draw=gray!15}, % --added, dashed, 
ylabel style={font=\color{white!15!black}},
ylabel={ERA metric},
axis background/.style={fill=white},
axis background/.style={fill=white} % --added
]
\addplot [color=mycolor1, mark=asterisk, mark options={solid, mycolor1}, thick]
  table[row sep=crcr]{%
1	0.02367297\\
2	0.03808953\\
3	0.04993541\\
4	0.05633177\\
};
%0.02367297 0.03808953 0.04993541 0.05633177
%\addlegendentry{RAND}

\addplot [color=mycolor2, mark=triangle, mark options={solid, mycolor2}, thick]
  table[row sep=crcr]{%
1	0.0236078\\
2	0.03749669\\
3	0.04976608\\
4	0.05639906\\
};
%0.0236078  0.03749669 0.04976608 0.05639906
%\addlegendentry{SEP}

\addplot [color=mycolor3, mark=o, mark options={solid, mycolor3}, thick]
  table[row sep=crcr]{%
1	0.02097653\\
2	0.03862419\\
3	0.05049124\\
4	0.06003639\\
};
%0.02097653 0.03862419 0.05049124 0.06003639
%\addlegendentry{DFE}

\addplot [color=mycolor4, mark=square*, mark options={solid, mycolor4}, thick]
  table[row sep=crcr]{%
1	0.05636127\\
2	0.10226411\\
3	0.1415649\\
4	0.17716534\\
};
%0.05636127 0.10226411 0.1415649  0.17716534
%\addlegendentry{DFE}

\addplot [color=mycolor5, mark=asterisk, mark options={solid, mycolor5}, thick]
  table[row sep=crcr]{%
1	0.05686425\\
2	0.10742434\\
3	0.15280348\\
4	0.1968154\\
};
%0.05686425 0.10742434 0.15280348 0.1968154

\addplot [color=mycolor6, mark=*, mark options={solid, mycolor6}, thick]
  table[row sep=crcr]{%
1	0.03263567\\
2	0.05478592\\
3	0.07699206\\
4	0.09252869\\
};
%0.03263567 0.05478592 0.07699206 0.09252869

\end{axis}
\end{tikzpicture}% 
            \caption[]%
            {{\small LSTM\_att / ERA}}
        \end{subfigure}
        \hfill
        \begin{subfigure}[b]{0.25\textwidth}  
            \setlength\figureheight{\heightunify in}
        \setlength\figurewidth{1.15in}
        \centering  \scriptsize
        % This file was created by matlab2tikz.
%
%The latest updates can be retrieved from
%  http://www.mathworks.com/matlabcentral/fileexchange/22022-matlab2tikz-matlab2tikz
%where you can also make suggestions and rate matlab2tikz.
%
\definecolor{mycolor1}{rgb}{1,0.65,0}%
\definecolor{mycolor2}{rgb}{1,0.94,0}%
\definecolor{mycolor3}{rgb}{1.0, 0.25, 0.25}%
\definecolor{mycolor4}{rgb}{0.39, 0.58, 0.93}
\definecolor{mycolor5}{rgb}{0., 0.18, 0.39}
\definecolor{mycolor6}{rgb}{0.01, 0.75, 0.24}

\begin{tikzpicture}

\begin{axis}[%
width=0.951\figurewidth,
height=\figureheight,
at={(0\figurewidth,0\figureheight)},
scale only axis,
scaled x ticks=true,
xticklabels={1,2,3,4},
xtick={1,2,3,4},
xmin=1,
xmax=4,
xlabel style={font=\color{white!15!black}},
xlabel={$s$},
ymin=0.0,
ymax=0.35,
grid, % --added
grid style={line width=.15pt, draw=gray!15}, % --added, dashed, 
ylabel style={font=\color{white!15!black}},
ylabel={MMA metric},
axis background/.style={fill=white},
axis background/.style={fill=white} % --added
]
\addplot [color=mycolor1, mark=asterisk, mark options={solid, mycolor1}, thick]
  table[row sep=crcr]{%
1	0.13159433\\
2	0.16242435\\
3	0.18023679\\
4	0.19555187\\
};
%0.13159433 0.16242435 0.18023679 0.19555187
%\addlegendentry{RAND}

\addplot [color=mycolor2, mark=triangle, mark options={solid, mycolor2}, thick]
  table[row sep=crcr]{%
1	0.13144139\\
2	0.16224503\\
3	0.18103833\\
4	0.19428186\\
};
%0.13144139 0.16224503 0.18103833 0.19428186
%\addlegendentry{SEP}

\addplot [color=mycolor3, mark=o, mark options={solid, mycolor3}, thick]
  table[row sep=crcr]{%
1	0.12105668\\
2	0.15862174\\
3	0.18192316\\
4	0.19892014\\
};
%0.12105668 0.15862174 0.18192316 0.19892014
%\addlegendentry{DFE}

\addplot [color=mycolor4, mark=square*, mark options={solid, mycolor4}, thick]
  table[row sep=crcr]{%
1	0.14073173\\
2	0.18272809\\
3	0.20566518\\
4	0.22272603\\
};
%0.14073173 0.18272809 0.20566518 0.22272603
%\addlegendentry{DFE}

\addplot [color=mycolor5, mark=asterisk, mark options={solid, mycolor5}, thick]
  table[row sep=crcr]{%
1	0.13361161\\
2	0.17884754\\
3	0.20858205\\
4	0.22604859\\
};
%0.13361161 0.17884754 0.20858205 0.22604859

\addplot [color=mycolor6, mark=*, mark options={solid, mycolor6}, thick]
  table[row sep=crcr]{%
1	0.14770942\\
2	0.21330508\\
3	0.25544182\\
4	0.28096764\\
};
%0.14770942 0.21330508 0.25544182 0.28096764

\end{axis}
\end{tikzpicture}% 
            \caption[]%
            {{\small LSTM\_att / MMA}}
        \end{subfigure}
        \hfill
        \caption[ The average and standard deviation of critical parameters ]
        {Interpretation faithfulness comparison under different metrics on AGNews dataset. \vagrad:VaGrad,\, \smoothgrad:SmoothGrad,\, \itergrad:IterGrad,\, \inpgrad:InpGrad,\, \integrad:InteGrad,\, \rankmask:RankMask.} 
        \label{fig:ag_cross}
        \vspace{-5pt}
\end{figure*}

The performance evaluation results on SST2 dataset are shown in Figure~\ref{fig:sst2_cross}. The results on Yelp and AGNews are in Figure~\ref{fig:yelp_cross} and Figure~\ref{fig:ag_cross}, respectively. Each result is averaged over $300$ instances randomly sampled from test data. In general, the trends of curves on different datasets are similar, so we only put SST2 results above. Results on Yelp and AGNews could be found at the end of the paper. Some observations could be made as below.
\begin{itemize}[leftmargin=*, topsep=0pt, noitemsep]
    \item First, an interpretation algorithm tends to have better performance if it is evaluated under the metric derived from the same definition. Specifically, when evaluated under the CSA metric, VaGrad, SmoothGrad and IterGrad have better performances than others. When evaluated under the ERA metric, InpGrad and InteGrad tend to perform better. When evaluated under the MMA metric, RankMask performs the best.
    \item Second, within each definition, the precision varies for different interpretation algorithms. In general, in CSA, IterGrad has the best performance since it is an iterative optimization algorithm. SmoothGrad is slightly better than VaGrad (although their curves are very close to each other in the figure), because the former neutralizes noise in prediction. In ERA, InteGrad has better performance than InpGrad, since the former computes feature contribution with finer granularity. To further validate this, we measure the performance of InteGrad by varying the number $T$ of points placed between $\textbf{x}$ and zero embeddings. The influences on InteGrad are shown in Figure~\ref{fig:integrad_sst2}. We only show results on SST2 for LSTM and BERT due to space limit (other results can be found in Appendix). The result shows that putting more points in the path improves accuracy of InteGrad. This trend is more significant in BERT than in LSTM, which indicates the prediction function in Bert is less smooth and sophisticated methods are needed to interpret it.
    \item Third, the superiority of different algorithms are relatively random if they are all mismatched to the metric. For examples, under CSA metric, InteGrad is \textit{not} consistently better than InpGrad. Also, under MMA metric, except RankMask, the performances of other methods is pretty close.
\end{itemize}

Therefore, from the analysis above, we could see that whether interpretation is ``faithful" largely depends on whether the evaluation metric is coupled with the algorithm producing interpretation. For example, RankMask does not seem to be as faithful as InpGrad and InteGrad when evaluated using $f_c(\textbf{x}_{-\mathcal{S}}) - f_c(\textbf{x}^*)$ as the metric. This is because, attention scores, by definition, do not represent the relation between input embeddings and output predictions. However, when using Equation~\ref{eq:mma} as the metric, the standard changes so that InpGrad and InteGrad are not accurate in understanding the message passing between latent representations.

The conclusion above provides a guideline of how to choose baseline methods in interpretation algorithm evaluation. A more rigorous setting is to first specify the definition, and then design the algorithms as well as choosing the metric, both based on the specified definition. If we want to compare between SmoothGrad (defined under CSA) and InteGrad (defined under ERA), then what matters is \textit{not} which algorithm is more ``faithful", but which specific evaluation metric is used. An interpretation algorithm is \textit{inherently more advantageous} when evaluated under the metric deriving from the same definition. Choosing the evaluation metric is rather a subjective task, depending on the application scenario or the preferences of audience, which could be a direction for future work.

\begin{figure}[t]
\centering
 \includegraphics[width=0.32\textwidth]{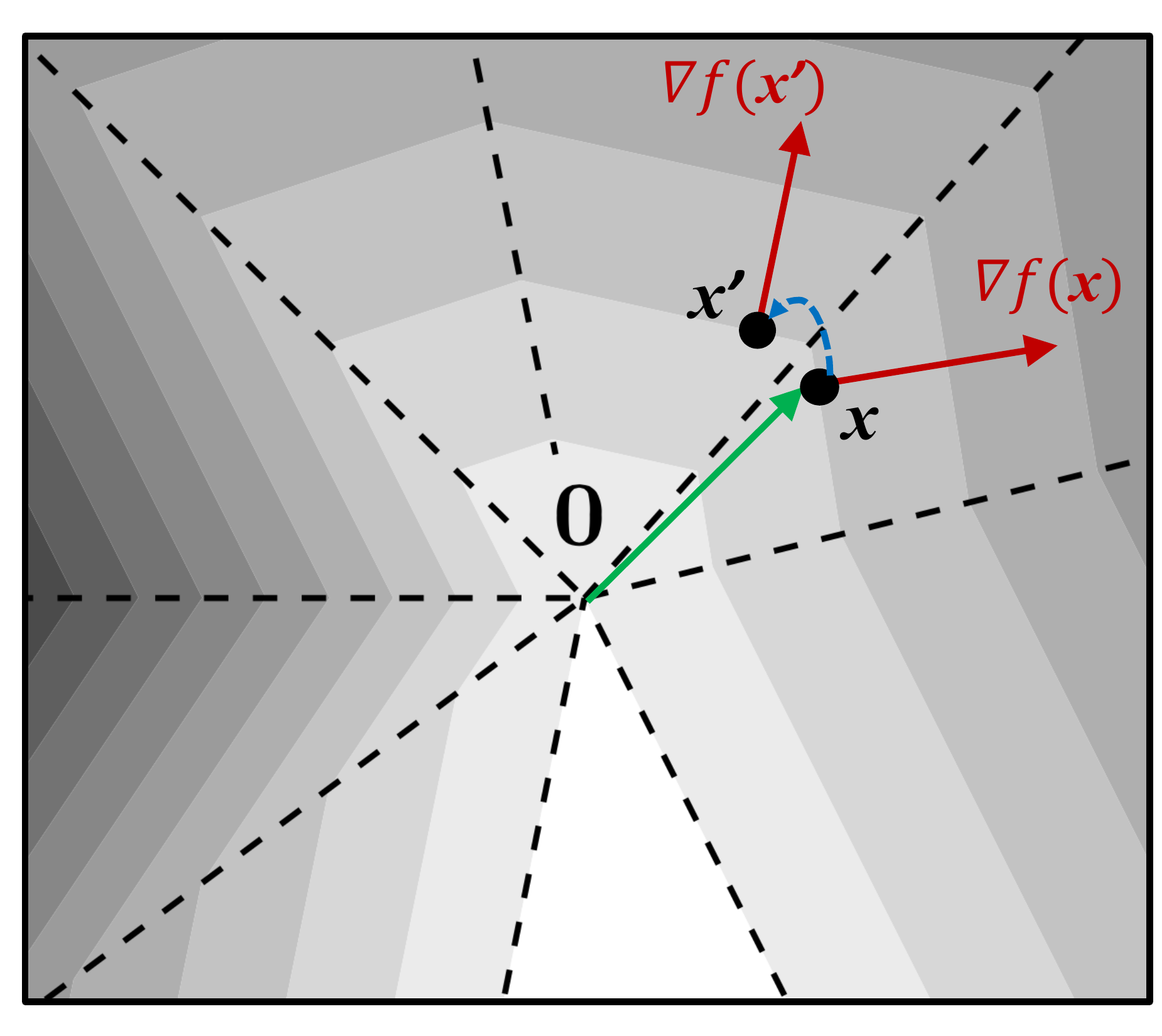}
 \caption{Interpretation analysis of a toy function.} \label{fig:toy}
 \vspace{-4pt}
\end{figure}

\subsection{Interpretation Properties Revisited}
Given the results at hand, we review the two properties discussed in~\cite{Jain-Wallace19attNotExplain, Wiegreffe-Pinter19attNotNotExplain} for faithful explanation. The first property is that a faithful interpretation should correlate with feature importance measures (e.g., gradient-based methods or erasure-based methods). The second property is that, changes to a faithful interpretation will cause prediction to vary.

The first property is already analyzed through the experiment in Section~\ref{sec:exp1_result}. Our theoretical and experimental analysis show that interpretations from different definitions have equal status, so we may not deny an interpretation's faithfulness due to its dissimilarity to another (or another group), even though the later is more commonly used or more intuitive. The metric for interpretation faithfulness is even not unique~\cite{haldar2019applying, Li-Jurafsky17erasure, du2018towards, hechtlinger2016interpretation}.

As for the second property, recent work shows that it is more relevant to the sensitivity of interpretation, rather than faithfulness~\cite{Ghorbani-etal19fragile, Dombrowski-etal19geometry}. The sensitivity could be influenced by characteristics (e.g., Hessian matrix) of prediction function. An illustrative example is in Figure~\ref{fig:toy}. The toy function's prediction values is marked with grey levels. By slightly perturbing $\textbf{x}$ to $\textbf{x}'$, we dramatically change its interpretation, although the prediction value does not change. However, it may be too rash to conclude that gradient-based interpretation (either VaGrad or InpGrad) does not faithfully reflect the rationale behind prediction. Therefore, it is hard to rule out function characteristics as the reason for the instability of interpretation. Future work could further explore how to balance different aspects (e.g., faithfulness, sensitivity, understandability) of interpretation given certain application scenarios.
%Recent work shows that this is more relevant to the sensitivity of interpretation~\cite{yeh2019fidelity} which is a different concept from faithfulness.

\begin{figure}[t]
        \centering
        \begin{subfigure}[b]{0.23\textwidth}
        \setlength\figureheight{0.80in}
        \setlength\figurewidth{1.15in}
        \centering  \scriptsize
        % This file was created by matlab2tikz.
%
%The latest updates can be retrieved from
%  http://www.mathworks.com/matlabcentral/fileexchange/22022-matlab2tikz-matlab2tikz
%where you can also make suggestions and rate matlab2tikz.
%
\definecolor{mycolor1}{rgb}{1,0.65,0}%
\definecolor{mycolor2}{rgb}{1,0.94,0}%
\definecolor{mycolor3}{rgb}{1.0, 0.25, 0.25}%
\definecolor{mycolor4}{rgb}{0.39, 0.58, 0.93}
\definecolor{mycolor5}{rgb}{0., 0.18, 0.39}

\begin{tikzpicture}

\begin{axis}[%
width=0.951\figurewidth,
height=\figureheight,
at={(0\figurewidth,0\figureheight)},
scale only axis,
scaled x ticks=true,
xticklabels={0,1,2,3,4},
xtick={1,2,3,4,5},
xmin=1,
xmax=5,
xlabel style={font=\color{white!15!black}},
xlabel={iteration},
ymin=0.15,
ymax=0.70,
grid, % --added
grid style={line width=.15pt, draw=gray!15}, % --added, dashed, 
ylabel style={font=\color{white!15!black}},
ylabel={similarity},
axis background/.style={fill=white},
legend columns = 2,
legend style={legend cell align=left, align=left, draw=white!15!black, nodes={scale=0.7}, at={(0.90, 0.97)}},
axis background/.style={fill=white} % --added
]
\addplot [color=mycolor5, mark=square*, mark options={solid, mycolor5}, thick]
  table[row sep=crcr]{%
1	0.2852\\
2	0.3556\\
3	0.4081\\
4	0.4221\\
5	0.4291\\
};
%0.17992776 0.32348302 0.40808931 0.46268828
\addlegendentry{retraining}

% \addplot [color=mycolor5, mark=triangle, mark options={solid, mycolor5}, thick]
%   table[row sep=crcr]{%
% 1	0.1969755\\
% 2	0.34623118\\
% 3	0.45489708\\
% 4	0.52438925\\
% };
% %0.1969755  0.34623118 0.45489708 0.52438925
% \addlegendentry{T=2}

\end{axis}
\end{tikzpicture}% 
            \caption[Network2]%
            {{\small SST2}}
        \end{subfigure}
        \hfill
        \begin{subfigure}[b]{0.23\textwidth}
        \setlength\figureheight{0.80in}
        \setlength\figurewidth{1.15in}
        \centering  \scriptsize
        % This file was created by matlab2tikz.
%
%The latest updates can be retrieved from
%  http://www.mathworks.com/matlabcentral/fileexchange/22022-matlab2tikz-matlab2tikz
%where you can also make suggestions and rate matlab2tikz.
%
\definecolor{mycolor1}{rgb}{1,0.65,0}%
\definecolor{mycolor2}{rgb}{1,0.94,0}%
\definecolor{mycolor3}{rgb}{1.0, 0.25, 0.25}%
\definecolor{mycolor4}{rgb}{0.39, 0.58, 0.93}
\definecolor{mycolor5}{rgb}{0., 0.18, 0.39}

\begin{tikzpicture}

\begin{axis}[%
width=0.951\figurewidth,
height=\figureheight,
at={(0\figurewidth,0\figureheight)},
scale only axis,
scaled x ticks=true,
xticklabels={0,1,2,3,4},
xtick={1,2,3,4,5},
xmin=1,
xmax=5,
xlabel style={font=\color{white!15!black}},
xlabel={iteration},
ymin=0.20,
ymax=0.70,
grid, % --added
grid style={line width=.15pt, draw=gray!15}, % --added, dashed, 
ylabel style={font=\color{white!15!black}},
ylabel={similarity},
axis background/.style={fill=white},
legend columns = 2,
legend style={legend cell align=left, align=left, draw=white!15!black, nodes={scale=0.7}, at={(0.90, 0.97)}},
axis background/.style={fill=white} % --added
]
\addplot [color=mycolor5, mark=square*, mark options={solid, mycolor5}, thick]
  table[row sep=crcr]{%
1	0.3821\\
2	0.4473\\
3	0.4782\\
4	0.4856\\
5	0.4912\\
};
%0.17992776 0.32348302 0.40808931 0.46268828
\addlegendentry{retraining}

% \addplot [color=mycolor5, mark=triangle, mark options={solid, mycolor5}, thick]
%   table[row sep=crcr]{%
% 1	0.1969755\\
% 2	0.34623118\\
% 3	0.45489708\\
% 4	0.52438925\\
% };
% %0.1969755  0.34623118 0.45489708 0.52438925
% \addlegendentry{T=2}

\end{axis}
\end{tikzpicture}% 
            \caption[Network2]%
            {{\small Yelp}}
        \end{subfigure}
        \caption[ The average and standard deviation of critical parameters ]
        {Alignment of interpretation to human cognition on LSTM.} 
        \label{fig:align_lstm}
        \vspace{-6pt}
\end{figure}
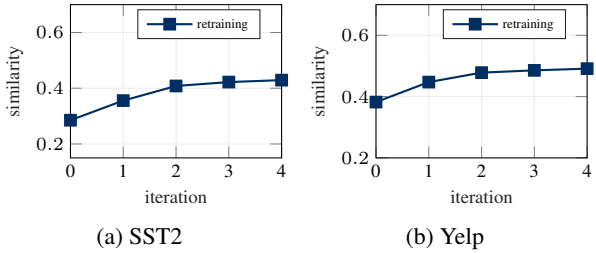

\subsection{Interpretation Alignment}
In this part, we evaluate how interpretation could be aligned with human cognition habits by applying the method introduced in Section~\ref{sec:align}.

For both SST2 and Yelp data, we manually sampled $100$ instances from the test set. In each sample, we select a number of important words as human knowledge $\mathcal{I}_{seed}$. The neighborhood size is chosen as $8$ for each embedding. We only involve SST2 and Yelp data because they are used in sentiment classification, and the important words are relatively easy to annotate manually. We choose CSA as the definition for interpretation, and let $\epsilon=0.5$. The importance score of word $n$ is computed as $s_n = \| \textbf{x}^n - {\textbf{x}^*}^n \|_2$. The similarity between interpretation and human cognition is defined as $\frac{\sum_{n\in \mathcal{I}} s_n }{\sum_{n} s_n}$. As we retrain the model, improved alignment on the LSTM model is shown in Figure~\ref{fig:align_lstm}. Interpretation becomes more similar to human cognition habits as retraining iterates. A side effect is that classification performance will slightly decrease. The resultant accuracy after retraining is $0.791$ for SST2 and $0.922$ for LSTM. Accuracy drops are common in adversarial training~\cite{Zhang-etal19theoretically}, which also explains this phenomenon.

\section{Related Work}

Various categories of interpretation methods have been proposed in recent years. Approaches for understanding importance of features in a given prediction include gradient based methods~\cite{Simonyan-etal13deepInsideCNNsaliency, hechtlinger2016interpretation, Smilkov-etal18smoothgrad, denil2014extraction, Sundararajan-etal16integratedGradient}, mimic learning based methods~\cite{Ribe-etal16why, che2016interpretable}, erasure based methods~\cite{Li-Jurafsky17erasure, Serrano-Smith19isAttIntp, Fong-Vedaldi17perturbation}, and a recent information theory based method~\cite{Guan-etal19unifiedNLP}. Approaches for understanding the high-level semantics of latent representations include~\cite{fyshe2015compositional, Kim-etal18concepts, Mathew-etal19polarInterpWordEmbd, Panigrahi-etal19word2sense}. In addition, \cite{Koh-Liang17influencefunction} proposes to understand importance of individual training samples as interpretation. As we have illustrated in this paper, many of the methods could be related and unified under the definition based on adversarial attack. 

Besides post-hoc interpretation, various work also propose to enable inherent interpretability in models. In general, these models rely on attention mechanisms~\cite{Bahdanau-etal15}. Some well known examples include transformer~\cite{vaswani2017attention}, capsule networks~\cite{Sabour-Hinton17routing}, graph convolutional networks~\cite{Kipf-Welling17semisupervised, Velickovic-etal18graphAttention}.

Besides developing specific algorithms to obtain interpretation, researchers have also initiated fundamental discussions over the topic. For examples, \cite{Lipt16themythos} discusses several desideratas of interpretability. \cite{Rudin-18pleasestop} discusses limitations of post-hoc interpretations. \cite{riedl2019human} analyzes the role of human in interpretable machine learning. \cite{Jain-Wallace19attNotExplain} discusses the potential limitation of attention in prediction interpretation.

\section{Conclusion and Future Work}
Through both theoretical analysis and experiments, we discover that significant bias may exist when evaluating interpretation methods, if baseline methods and the metric are not carefully chosen. To avoid false intuition on interpretation faithfulness, we propose a definition driven pipeline to guide the development and evaluation of interpretation methods. We use adversarial attack as the general definition, and show its equivalence to various existing interpretation methods. After that, we propose a new method to promote model interpretability when human assessment is involved.

Future work include considering more applications to be interpreted, such as question answering and natural language inference. Besides, it remains unsolved how to avoid affecting model performance while incorporating human cognition.

\newpage
\bibliographystyle{acl_natbib}
\bibliography{emnlp2020}

\newpage~\newpage
\section{Appendix}
\begin{figure}[h]
        \centering
        \begin{subfigure}[b]{0.28\textwidth}
        \setlength\figureheight{1.20in}
        \setlength\figurewidth{1.35in}
        \centering  \scriptsize
        % This file was created by matlab2tikz.
%
%The latest updates can be retrieved from
%  http://www.mathworks.com/matlabcentral/fileexchange/22022-matlab2tikz-matlab2tikz
%where you can also make suggestions and rate matlab2tikz.
%
\definecolor{mycolor1}{rgb}{1,0.65,0}%
\definecolor{mycolor2}{rgb}{0.39, 0.58, 0.93}%
\definecolor{mycolor3}{rgb}{0.0, 0.8, 0.6}%
\definecolor{mycolor4}{rgb}{0.37, 0.62, 0.63}
\definecolor{mycolor5}{rgb}{0., 0.18, 0.39}

\begin{tikzpicture}

\begin{axis}[%
width=0.951\figurewidth,
height=\figureheight,
at={(0\figurewidth,0\figureheight)},
scale only axis,
scaled x ticks=true,
xticklabels={1,2,3,4},
xtick={1,2,3,4},
xmin=1,
xmax=4,
xlabel style={font=\color{white!15!black}},
xlabel={$s$},
ymin=0.15,
ymax=0.70,
grid, % --added
grid style={line width=.15pt, draw=gray!15}, % --added, dashed, 
ylabel style={font=\color{white!15!black}},
ylabel={ERA metric},
axis background/.style={fill=white},
legend columns = 2,
legend style={legend cell align=left, align=left, draw=white!15!black, nodes={scale=0.7}, at={(0.90, 0.97)}},
axis background/.style={fill=white} % --added
]
\addplot [color=mycolor2, mark=square*, mark options={solid, mycolor2}, thick]
  table[row sep=crcr]{%
1	0.17992776\\
2	0.32348302\\
3	0.40808931\\
4	0.46268828\\
};
%0.17992776 0.32348302 0.40808931 0.46268828
\addlegendentry{T=1}

\addplot [color=mycolor3, mark=triangle, mark options={solid, mycolor3}, thick]
  table[row sep=crcr]{%
1	0.1969755\\
2	0.34623118\\
3	0.45489708\\
4	0.52438925\\
};
%0.1969755  0.34623118 0.45489708 0.52438925
\addlegendentry{T=2}

\addplot [color=mycolor4, mark=o, mark options={solid, mycolor4}, thick]
  table[row sep=crcr]{%
1	0.19930865\\
2	0.35388363\\
3	0.46674313\\
4	0.53965465\\
};
%0.19930865 0.35388363 0.46674313 0.53965465
\addlegendentry{T=4}

\addplot [color=mycolor5, mark=asterisk, mark options={solid, mycolor5}, thick]
  table[row sep=crcr]{%
1	0.19807485\\
2	0.35524494\\
3	0.46730655\\
4	0.54360266\\
};
%0.19807485 0.35524494 0.46730655 0.54360266
\addlegendentry{T=8}

\end{axis}
\end{tikzpicture}% 
            \caption[Network2]%
            {{\small LSTM / ERA}}
        \end{subfigure}
        %\hfill
        \hspace{10pt}
        \begin{subfigure}[b]{0.28\textwidth}
        \setlength\figureheight{1.20in}
        \setlength\figurewidth{1.35in}
        \centering  \scriptsize
        % This file was created by matlab2tikz.
%
%The latest updates can be retrieved from
%  http://www.mathworks.com/matlabcentral/fileexchange/22022-matlab2tikz-matlab2tikz
%where you can also make suggestions and rate matlab2tikz.
%
\definecolor{mycolor1}{rgb}{1,0.65,0}%
\definecolor{mycolor2}{rgb}{0.39, 0.58, 0.93}%
\definecolor{mycolor3}{rgb}{0.0, 0.8, 0.6}%
\definecolor{mycolor4}{rgb}{0.37, 0.62, 0.63}
\definecolor{mycolor5}{rgb}{0., 0.18, 0.39}

\begin{tikzpicture}

\begin{axis}[%
width=0.951\figurewidth,
height=\figureheight,
at={(0\figurewidth,0\figureheight)},
scale only axis,
scaled x ticks=true,
xticklabels={1,2,3,4},
xtick={1,2,3,4},
xmin=1,
xmax=4,
xlabel style={font=\color{white!15!black}},
xlabel={$s$},
ymin=0.00,
ymax=0.65,
grid, % --added
grid style={line width=.15pt, draw=gray!15}, % --added, dashed, 
ylabel style={font=\color{white!15!black}},
ylabel={ERA metric},
axis background/.style={fill=white},
legend columns = 2,
legend style={legend cell align=left, align=left, draw=white!15!black, nodes={scale=0.7}, at={(0.95, 0.97)}},
axis background/.style={fill=white} % --added
]
\addplot [color=mycolor2, mark=square*, mark options={solid, mycolor2}, thick]
  table[row sep=crcr]{%
1	0.08384714\\
2	0.11675363\\
3	0.14601074\\
4	0.18890988\\
};
%0.08384714 0.11675363 0.14601074 0.18890988
\addlegendentry{T=1}

\addplot [color=mycolor3, mark=triangle, mark options={solid, mycolor3}, thick]
  table[row sep=crcr]{%
1	0.16746232\\
2	0.27369746\\
3	0.33361747\\
4	0.39662887\\
};
%0.16746232 0.27369746 0.33361747 0.39662887
\addlegendentry{T=5}

\addplot [color=mycolor4, mark=o, mark options={solid, mycolor4}, thick]
  table[row sep=crcr]{%
1	0.16540853\\
2	0.30416488\\
3	0.36892283\\
4	0.43001318\\
};
%0.16540853 0.30416488 0.36892283 0.43001318
\addlegendentry{T=10}

\addplot [color=mycolor5, mark=asterisk, mark options={solid, mycolor5}, thick]
  table[row sep=crcr]{%
1	0.17686895\\
2	0.32740351\\
3	0.39470266\\
4	0.45144566\\
};
%0.17686895 0.32740351 0.39470266 0.45144566
\addlegendentry{T=20}

\end{axis}
\end{tikzpicture}% 
            \caption[Network2]%
            {{\small BERT / ERA}}
        \end{subfigure}
        %\hfill
        \hspace{10pt}
        \begin{subfigure}[b]{0.28\textwidth}
        \setlength\figureheight{1.20in}
        \setlength\figurewidth{1.35in}
        \centering  \scriptsize
        % This file was created by matlab2tikz.
%
%The latest updates can be retrieved from
%  http://www.mathworks.com/matlabcentral/fileexchange/22022-matlab2tikz-matlab2tikz
%where you can also make suggestions and rate matlab2tikz.
%
\definecolor{mycolor1}{rgb}{1,0.65,0}%
\definecolor{mycolor2}{rgb}{0.39, 0.58, 0.93}%
\definecolor{mycolor3}{rgb}{0.0, 0.8, 0.6}%
\definecolor{mycolor4}{rgb}{0.37, 0.62, 0.63}
\definecolor{mycolor5}{rgb}{0., 0.18, 0.39}

\begin{tikzpicture}

\begin{axis}[%
width=0.951\figurewidth,
height=\figureheight,
at={(0\figurewidth,0\figureheight)},
scale only axis,
scaled x ticks=true,
xticklabels={1,2,3,4},
xtick={1,2,3,4},
xmin=1,
xmax=4,
xlabel style={font=\color{white!15!black}},
xlabel={$s$},
ymin=0.20,
ymax=0.75,
grid, % --added
grid style={line width=.15pt, draw=gray!15}, % --added, dashed, 
ylabel style={font=\color{white!15!black}},
ylabel={ERA metric},
axis background/.style={fill=white},
legend columns = 2,
legend style={legend cell align=left, align=left, draw=white!15!black, nodes={scale=0.7}, at={(0.95, 0.97)}},
axis background/.style={fill=white} % --added
]
\addplot [color=mycolor2, mark=square*, mark options={solid, mycolor2}, thick]
  table[row sep=crcr]{%
1	0.23334137\\
2	0.39715452\\
3	0.5098031\\
4	0.57981006\\
};
%0.23334137 0.39715452 0.5098031  0.57981006
\addlegendentry{T=1}

\addplot [color=mycolor3, mark=triangle, mark options={solid, mycolor3}, thick]
  table[row sep=crcr]{%
1	0.2420149\\
2	0.41576346\\
3	0.53305202\\
4	0.60710714\\
};
%0.2420149  0.41576346 0.53305202 0.60710714
\addlegendentry{T=2}

\addplot [color=mycolor4, mark=o, mark options={solid, mycolor4}, thick]
  table[row sep=crcr]{%
1	0.24289394\\
2	0.41938002\\
3	0.5362582\\
4	0.61524113\\
};
%0.24289394 0.41938002 0.5362582  0.61524113
\addlegendentry{T=4}

\addplot [color=mycolor5, mark=asterisk, mark options={solid, mycolor5}, thick]
  table[row sep=crcr]{%
1	0.24326848\\
2	0.41994287\\
3	0.53790173\\
4	0.61632277\\
};
%0.24326848 0.41994287 0.53790173 0.61632277
\addlegendentry{T=8}

\end{axis}
\end{tikzpicture}% 
            \caption[Network2]%
            {{\small LSTM\_att / ERA}}
        \end{subfigure}
        \caption[ The average and standard deviation of critical parameters ]
        {Parameter analysis of InteGrad on SST2.} 
        \label{fig:integrad_sst2_}
        \vspace{-4pt}
\end{figure}
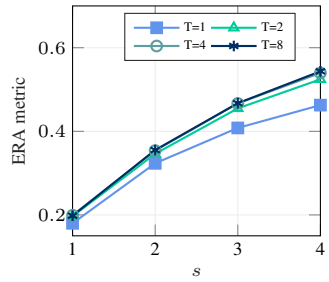
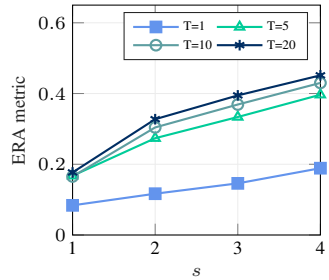
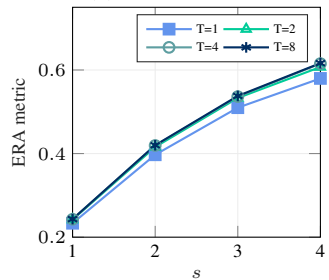

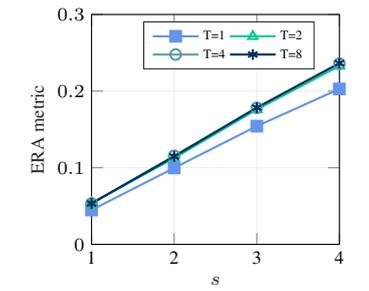
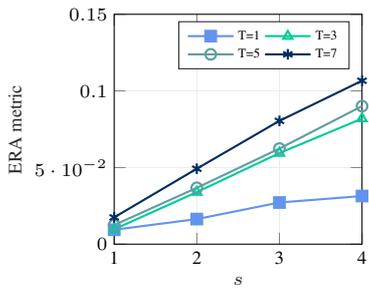
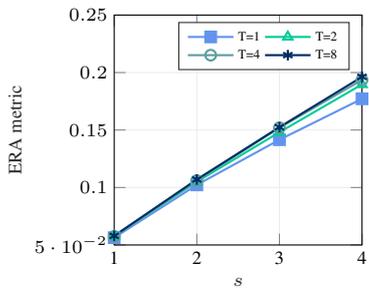
\begin{figure}[t]
        \centering
        \begin{subfigure}[b]{0.32\textwidth}
        \setlength\figureheight{1.20in}
        \setlength\figurewidth{1.35in}
        \centering  \scriptsize
        % This file was created by matlab2tikz.
%
%The latest updates can be retrieved from
%  http://www.mathworks.com/matlabcentral/fileexchange/22022-matlab2tikz-matlab2tikz
%where you can also make suggestions and rate matlab2tikz.
%
\definecolor{mycolor1}{rgb}{1,0.65,0}%
\definecolor{mycolor2}{rgb}{0.39, 0.58, 0.93}%
\definecolor{mycolor3}{rgb}{0.0, 0.8, 0.6}%
\definecolor{mycolor4}{rgb}{0.37, 0.62, 0.63}
\definecolor{mycolor5}{rgb}{0., 0.18, 0.39}

\begin{tikzpicture}

\begin{axis}[%
width=0.951\figurewidth,
height=\figureheight,
at={(0\figurewidth,0\figureheight)},
scale only axis,
scaled x ticks=true,
xticklabels={1,2,3,4},
xtick={1,2,3,4},
xmin=1,
xmax=4,
xlabel style={font=\color{white!15!black}},
xlabel={$s$},
ymin=0.00,
ymax=0.30,
grid, % --added
grid style={line width=.15pt, draw=gray!15}, % --added, dashed, 
ylabel style={font=\color{white!15!black}},
ylabel={ERA metric},
axis background/.style={fill=white},
legend columns = 2,
legend style={legend cell align=left, align=left, draw=white!15!black, nodes={scale=0.7}, at={(0.90, 0.97)}},
axis background/.style={fill=white} % --added
]
\addplot [color=mycolor2, mark=square*, mark options={solid, mycolor2}, thick]
  table[row sep=crcr]{%
1	0.04481789\\
2	0.09971746\\
3	0.15429465\\
4	0.20280115\\
};
%0.04481789 0.09971746 0.15429465 0.20280115
\addlegendentry{T=1}

\addplot [color=mycolor3, mark=triangle, mark options={solid, mycolor3}, thick]
  table[row sep=crcr]{%
1	0.05377946\\
2	0.11257111\\
3	0.17598104\\
4	0.23273849\\
};
%0.05377946 0.11257111 0.17598104 0.23273849
\addlegendentry{T=2}

\addplot [color=mycolor4, mark=o, mark options={solid, mycolor4}, thick]
  table[row sep=crcr]{%
1	0.05350378\\
2	0.11564279\\
3	0.1781747\\
4	0.23578354\\
};
%0.05350378 0.11564279 0.1781747  0.23578354
\addlegendentry{T=4}

\addplot [color=mycolor5, mark=asterisk, mark options={solid, mycolor5}, thick]
  table[row sep=crcr]{%
1	0.05339326\\
2	0.11468772\\
3	0.178396\\
4	0.23643169\\
};
%0.05339326 0.11468772 0.178396   0.23643169
\addlegendentry{T=8}

\end{axis}
\end{tikzpicture}% 
            \caption[Network2]%
            {{\small LSTM / ERA}}
        \end{subfigure}
        %\hfill
        \hspace{0pt}
        \begin{subfigure}[b]{0.32\textwidth}
        \setlength\figureheight{1.20in}
        \setlength\figurewidth{1.35in}
        \centering  \scriptsize
        % This file was created by matlab2tikz.
%
%The latest updates can be retrieved from
%  http://www.mathworks.com/matlabcentral/fileexchange/22022-matlab2tikz-matlab2tikz
%where you can also make suggestions and rate matlab2tikz.
%
\definecolor{mycolor1}{rgb}{1,0.65,0}%
\definecolor{mycolor2}{rgb}{0.39, 0.58, 0.93}%
\definecolor{mycolor3}{rgb}{0.0, 0.8, 0.6}%
\definecolor{mycolor4}{rgb}{0.37, 0.62, 0.63}
\definecolor{mycolor5}{rgb}{0., 0.18, 0.39}

\begin{tikzpicture}

\begin{axis}[%
width=0.951\figurewidth,
height=\figureheight,
at={(0\figurewidth,0\figureheight)},
scale only axis,
scaled x ticks=true,
xticklabels={1,2,3,4},
xtick={1,2,3,4},
xmin=1,
xmax=4,
xlabel style={font=\color{white!15!black}},
xlabel={$s$},
ymin=0.00,
ymax=0.15,
grid, % --added
grid style={line width=.15pt, draw=gray!15}, % --added, dashed, 
ylabel style={font=\color{white!15!black}},
ylabel={ERA metric},
axis background/.style={fill=white},
legend columns = 2,
legend style={legend cell align=left, align=left, draw=white!15!black, nodes={scale=0.7}, at={(0.95, 0.97)}},
axis background/.style={fill=white} % --added
]
\addplot [color=mycolor2, mark=square*, mark options={solid, mycolor2}, thick]
  table[row sep=crcr]{%
1	0.00939508\\
2	0.01637094\\
3	0.02716969\\
4	0.03140361\\
};
%0.00939508 0.01637094 0.02716969 0.03140361
\addlegendentry{T=1}

\addplot [color=mycolor3, mark=triangle, mark options={solid, mycolor3}, thick]
  table[row sep=crcr]{%
1	0.00970148\\
2	0.03395822\\
3	0.05940095\\
4	0.08194924\\
};
%0.00970148 0.03395822 0.05940095 0.08194924
\addlegendentry{T=3}

\addplot [color=mycolor4, mark=o, mark options={solid, mycolor4}, thick]
  table[row sep=crcr]{%
1	0.01246302\\
2	0.03676204\\
3	0.06237878\\
4	0.09005652\\
};
%0.01246302 0.03676204 0.06237878 0.09005652
\addlegendentry{T=5}

\addplot [color=mycolor5, mark=asterisk, mark options={solid, mycolor5}, thick]
  table[row sep=crcr]{%
1	0.01747966\\
2	0.04922929\\
3	0.08046139\\
4	0.10664239\\
};
%0.01747966 0.04922929 0.08046139 0.10664239
\addlegendentry{T=7}

\end{axis}
\end{tikzpicture}% 
            \caption[Network2]%
            {{\small BERT / ERA}}
        \end{subfigure}
        %\hfill
        \hspace{0pt}
        \begin{subfigure}[b]{0.32\textwidth}
        \setlength\figureheight{1.20in}
        \setlength\figurewidth{1.35in}
        \centering  \scriptsize
        % This file was created by matlab2tikz.
%
%The latest updates can be retrieved from
%  http://www.mathworks.com/matlabcentral/fileexchange/22022-matlab2tikz-matlab2tikz
%where you can also make suggestions and rate matlab2tikz.
%
\definecolor{mycolor1}{rgb}{1,0.65,0}%
\definecolor{mycolor2}{rgb}{0.39, 0.58, 0.93}%
\definecolor{mycolor3}{rgb}{0.0, 0.8, 0.6}%
\definecolor{mycolor4}{rgb}{0.37, 0.62, 0.63}
\definecolor{mycolor5}{rgb}{0., 0.18, 0.39}

\begin{tikzpicture}

\begin{axis}[%
width=0.951\figurewidth,
height=\figureheight,
at={(0\figurewidth,0\figureheight)},
scale only axis,
scaled x ticks=true,
xticklabels={1,2,3,4},
xtick={1,2,3,4},
xmin=1,
xmax=4,
xlabel style={font=\color{white!15!black}},
xlabel={$s$},
ymin=0.05,
ymax=0.25,
grid, % --added
grid style={line width=.15pt, draw=gray!15}, % --added, dashed, 
ylabel style={font=\color{white!15!black}},
ylabel={ERA metric},
axis background/.style={fill=white},
legend columns = 2,
legend style={legend cell align=left, align=left, draw=white!15!black, nodes={scale=0.7}, at={(0.95, 0.97)}},
axis background/.style={fill=white} % --added
]
\addplot [color=mycolor2, mark=square*, mark options={solid, mycolor2}, thick]
  table[row sep=crcr]{%
1	0.05636127\\
2	0.10226411\\
3	0.1415649\\
4	0.17716534\\
};
%0.05636127 0.10226411 0.1415649  0.17716534
\addlegendentry{T=1}

\addplot [color=mycolor3, mark=triangle, mark options={solid, mycolor3}, thick]
  table[row sep=crcr]{%
1	0.05767466\\
2	0.10500725\\
3	0.14812756\\
4	0.18968695\\
};
%0.05767466 0.10500725 0.14812756 0.18968695
\addlegendentry{T=2}

\addplot [color=mycolor4, mark=o, mark options={solid, mycolor4}, thick]
  table[row sep=crcr]{%
1	0.05754901\\
2	0.10602936\\
3	0.15198003\\
4	0.19361154\\
};
%0.05754901 0.10602936 0.15198003 0.19361154
\addlegendentry{T=4}

\addplot [color=mycolor5, mark=asterisk, mark options={solid, mycolor5}, thick]
  table[row sep=crcr]{%
1	0.05794665\\
2	0.10696466\\
3	0.15249039\\
4	0.19625063\\
};
%0.05794665 0.10696466 0.15249039 0.19625063
\addlegendentry{T=8}

\end{axis}
\end{tikzpicture}% 
            \caption[Network2]%
            {{\small LSTM\_att / ERA}}
        \end{subfigure}
        \caption[ The average and standard deviation of critical parameters ]
        {Parameter analysis of InteGrad on AGNews.} 
        \label{fig:integrad_ag}
        \vspace{-4pt}
\end{figure}

In the appendix, we provide the full result of Integrated Gradient on all datasets from Figure~\ref{fig:integrad_sst2_}$\sim$ Figure~\ref{fig:integrad_ag}. Each plots show the performance changes of Integrated Gradient by varying the number of points $T$ between zero point and input point. This is to complement the results shown in Figure~\ref{fig:integrad_sst2} due to page limit. In general, we could observe that the performance of Integrated Gradient improves as more points are placed used in computation.

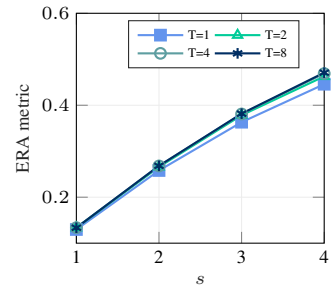
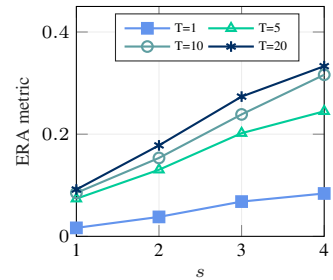
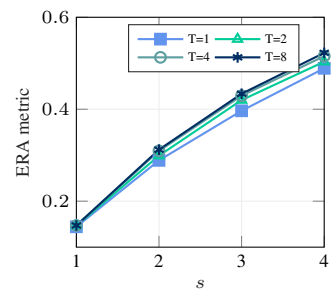
\begin{figure}[t]
        \centering
        \begin{subfigure}[b]{0.28\textwidth}
        \setlength\figureheight{1.20in}
        \setlength\figurewidth{1.35in}
        \centering  \scriptsize
        % This file was created by matlab2tikz.
%
%The latest updates can be retrieved from
%  http://www.mathworks.com/matlabcentral/fileexchange/22022-matlab2tikz-matlab2tikz
%where you can also make suggestions and rate matlab2tikz.
%
\definecolor{mycolor1}{rgb}{1,0.65,0}%
\definecolor{mycolor2}{rgb}{0.39, 0.58, 0.93}%
\definecolor{mycolor3}{rgb}{0.0, 0.8, 0.6}%
\definecolor{mycolor4}{rgb}{0.37, 0.62, 0.63}
\definecolor{mycolor5}{rgb}{0., 0.18, 0.39}

\begin{tikzpicture}

\begin{axis}[%
width=0.951\figurewidth,
height=\figureheight,
at={(0\figurewidth,0\figureheight)},
scale only axis,
scaled x ticks=true,
xticklabels={1,2,3,4},
xtick={1,2,3,4},
xmin=1,
xmax=4,
xlabel style={font=\color{white!15!black}},
xlabel={$s$},
ymin=0.10,
ymax=0.60,
grid, % --added
grid style={line width=.15pt, draw=gray!15}, % --added, dashed, 
ylabel style={font=\color{white!15!black}},
ylabel={ERA metric},
axis background/.style={fill=white},
legend columns = 2,
legend style={legend cell align=left, align=left, draw=white!15!black, nodes={scale=0.7}, at={(0.90, 0.97)}},
axis background/.style={fill=white} % --added
]
\addplot [color=mycolor2, mark=square*, mark options={solid, mycolor2}, thick]
  table[row sep=crcr]{%
1	0.12976001\\
2	0.25809675\\
3	0.36313524\\
4	0.44577235\\
};
%0.12976001 0.25809675 0.36313524 0.44577235
\addlegendentry{T=1}

\addplot [color=mycolor3, mark=triangle, mark options={solid, mycolor3}, thick]
  table[row sep=crcr]{%
1	0.13349935\\
2	0.2658962\\
3	0.37831093\\
4	0.46277422\\
};
%0.13349935 0.2658962  0.37831093 0.46277422
\addlegendentry{T=2}

\addplot [color=mycolor4, mark=o, mark options={solid, mycolor4}, thick]
  table[row sep=crcr]{%
1	0.13376777\\
2	0.26796456\\
3	0.38104328\\
4	0.46897536\\
};
%0.13376777 0.26796456 0.38104328 0.46897536
\addlegendentry{T=4}

\addplot [color=mycolor5, mark=asterisk, mark options={solid, mycolor5}, thick]
  table[row sep=crcr]{%
1	0.13362201\\
2	0.26821098\\
3	0.3817218\\
4	0.47089104\\
};
%0.13362201 0.26821098 0.3817218  0.47089104
\addlegendentry{T=8}

\end{axis}
\end{tikzpicture}% 
            \caption[Network2]%
            {{\small LSTM / ERA}}
        \end{subfigure}
        %\hfill
        \hspace{10pt}
        \begin{subfigure}[b]{0.28\textwidth}
        \setlength\figureheight{1.20in}
        \setlength\figurewidth{1.35in}
        \centering  \scriptsize
        % This file was created by matlab2tikz.
%
%The latest updates can be retrieved from
%  http://www.mathworks.com/matlabcentral/fileexchange/22022-matlab2tikz-matlab2tikz
%where you can also make suggestions and rate matlab2tikz.
%
\definecolor{mycolor1}{rgb}{1,0.65,0}%
\definecolor{mycolor2}{rgb}{0.39, 0.58, 0.93}%
\definecolor{mycolor3}{rgb}{0.0, 0.8, 0.6}%
\definecolor{mycolor4}{rgb}{0.37, 0.62, 0.63}
\definecolor{mycolor5}{rgb}{0., 0.18, 0.39}

\begin{tikzpicture}

\begin{axis}[%
width=0.951\figurewidth,
height=\figureheight,
at={(0\figurewidth,0\figureheight)},
scale only axis,
scaled x ticks=true,
xticklabels={1,2,3,4},
xtick={1,2,3,4},
xmin=1,
xmax=4,
xlabel style={font=\color{white!15!black}},
xlabel={$s$},
ymin=0.00,
ymax=0.45,
grid, % --added
grid style={line width=.15pt, draw=gray!15}, % --added, dashed, 
ylabel style={font=\color{white!15!black}},
ylabel={ERA metric},
axis background/.style={fill=white},
legend columns = 2,
legend style={legend cell align=left, align=left, draw=white!15!black, nodes={scale=0.7}, at={(0.90, 0.97)}},
axis background/.style={fill=white} % --added
]
\addplot [color=mycolor2, mark=square*, mark options={solid, mycolor2}, thick]
  table[row sep=crcr]{%
1	0.016594\\
2	0.03806318\\
3	0.06805319\\
4	0.08392287\\
};
%0.016594   0.03806318 0.06805319 0.08392287
\addlegendentry{T=1}

\addplot [color=mycolor3, mark=triangle, mark options={solid, mycolor3}, thick]
  table[row sep=crcr]{%
1	0.073684\\
2	0.13036031\\
3	0.20195388\\
4	0.24477567\\
};
%0.073684   0.13036031 0.20195388 0.24477567
\addlegendentry{T=5}

\addplot [color=mycolor4, mark=o, mark options={solid, mycolor4}, thick]
  table[row sep=crcr]{%
1	0.08507936\\
2	0.15346998\\
3	0.23870122\\
4	0.31630171\\
};
%0.08507936 0.15346998 0.23870122 0.31630171
\addlegendentry{T=10}

\addplot [color=mycolor5, mark=asterisk, mark options={solid, mycolor5}, thick]
  table[row sep=crcr]{%
1	0.09168428\\
2	0.17803603\\
3	0.27358104\\
4	0.33309593\\
};
%0.09168428 0.17803603 0.27358104 0.33309593
\addlegendentry{T=20}

\end{axis}
\end{tikzpicture}% 
            \caption[Network2]%
            {{\small BERT / ERA}}
        \end{subfigure}
        %\hfill
        \hspace{10pt}
        \begin{subfigure}[b]{0.28\textwidth}
        \setlength\figureheight{1.20in}
        \setlength\figurewidth{1.35in}
        \centering  \scriptsize
        % This file was created by matlab2tikz.
%
%The latest updates can be retrieved from
%  http://www.mathworks.com/matlabcentral/fileexchange/22022-matlab2tikz-matlab2tikz
%where you can also make suggestions and rate matlab2tikz.
%
\definecolor{mycolor1}{rgb}{1,0.65,0}%
\definecolor{mycolor2}{rgb}{0.39, 0.58, 0.93}%
\definecolor{mycolor3}{rgb}{0.0, 0.8, 0.6}%
\definecolor{mycolor4}{rgb}{0.37, 0.62, 0.63}
\definecolor{mycolor5}{rgb}{0., 0.18, 0.39}

\begin{tikzpicture}

\begin{axis}[%
width=0.951\figurewidth,
height=\figureheight,
at={(0\figurewidth,0\figureheight)},
scale only axis,
scaled x ticks=true,
xticklabels={1,2,3,4},
xtick={1,2,3,4},
xmin=1,
xmax=4,
xlabel style={font=\color{white!15!black}},
xlabel={$s$},
ymin=0.10,
ymax=0.60,
grid, % --added
grid style={line width=.15pt, draw=gray!15}, % --added, dashed, 
ylabel style={font=\color{white!15!black}},
ylabel={ERA metric},
axis background/.style={fill=white},
legend columns = 2,
legend style={legend cell align=left, align=left, draw=white!15!black, nodes={scale=0.7}, at={(0.90, 0.97)}},
axis background/.style={fill=white} % --added
]
\addplot [color=mycolor2, mark=square*, mark options={solid, mycolor2}, thick]
  table[row sep=crcr]{%
1	0.14430331\\
2	0.28875522\\
3	0.39662543\\
4	0.48892653\\
};
%0.14430331 0.28875522 0.39662543 0.48892653
\addlegendentry{T=1}

\addplot [color=mycolor3, mark=triangle, mark options={solid, mycolor3}, thick]
  table[row sep=crcr]{%
1	0.14650965\\
2	0.29900434\\
3	0.41984666\\
4	0.50334261\\
};
%0.14650965 0.29900434 0.41984666 0.50334261
\addlegendentry{T=2}

\addplot [color=mycolor4, mark=o, mark options={solid, mycolor4}, thick]
  table[row sep=crcr]{%
1	0.14710973\\
2	0.30937053\\
3	0.42943316\\
4	0.51546423\\
};
%0.14710973 0.30937053 0.42943316 0.51546423
\addlegendentry{T=4}

\addplot [color=mycolor5, mark=asterisk, mark options={solid, mycolor5}, thick]
  table[row sep=crcr]{%
1	0.14716823\\
2	0.31212834\\
3	0.4336269\\
4	0.52257752\\
};
%0.14716823 0.31212834 0.4336269  0.52257752
\addlegendentry{T=8}

\end{axis}
\end{tikzpicture}% 
            \caption[Network2]%
            {{\small LSTM\_att / ERA}}
        \end{subfigure}
        \caption[ The average and standard deviation of critical parameters ]
        {Parameter analysis of InteGrad on Yelp.} 
        \label{fig:integrad_yelp}
        \vspace{-4pt}
\end{figure}
~

\end{document}